\documentclass{article}
\usepackage{makecell}

\usepackage{arxiv}
\usepackage{float} 
\usepackage{longtable}
\usepackage{caption}
\usepackage{siunitx}
\usepackage[utf8]{inputenc} 
\usepackage[T1]{fontenc}    
\usepackage{hyperref}       
\usepackage{url}            
\usepackage{booktabs}       
\usepackage{amsfonts}       
\usepackage{nicefrac}       
\usepackage{microtype}      
\usepackage{lipsum}		
\usepackage{graphicx}
\usepackage[numbers]{natbib}
\usepackage{doi}
\usepackage[acronym]{glossaries}
\newacronym{cnn}{CNN}{Convolutional Neural Network}
\newacronym{vit}{ViT}{Vision Transformer}
\newacronym{nlp}{NLP}{Natural Language Processing}
\newacronym{rnn}{RNN}{Recurrent Neural Network}
\newacronym{dl}{DL}{Deep Learning}
\newacronym{ml}{ML}{Machine Learning}
\newacronym{mlp}{MLP}{Multi-Layer Perceptron}
\newacronym{gan}{GAN}{Generative Adversarial Network}
\newacronym{swint}{SwinT}{Swin Transformer}
\newacronym{cv}{CV}{Computer Vision}
\title{Vision Transformers in Precision Agriculture: A Comprehensive Survey}

\author{{\hspace{1mm}Saber Mehdipour} \\
	Department of Computer Engineering\\
	University of Guilan\\
	Rasht, Iran \\
	\texttt{sabermehdipour@webmail.guilan.ac.ir} \\
	\And
	{\hspace{1mm}Seyed Abolghasem Mirroshandel} \\
	Department of Computer Engineering\\
	University of Guilan\\
	Rasht, Iran \\
	\texttt{mirroshandel@guilan.ac.ir} \\
    \And
    {\hspace{1mm}Seyed Amirhossein Tabatabaei} \\
	Department of Computer Science\\
	University of Guilan\\
	Rasht, Iran \\
	\texttt{amirhossein.tabatabaei@guilan.ac.ir} \\
}

\date{}

\renewcommand{\headeright}{S.Mehdipour et al.}
\renewcommand{\shorttitle}{}

\hypersetup{
pdftitle={Vision Transformers in Precision Agriculture: A Comprehensive Survey},
pdfsubject={cs.CV, cs.AI},
pdfauthor={Saber Mehdipour, Seyed Abolghasem Mirroshandel, Seyed Amirhossein Tabatabaei},
pdfkeywords={Vision Transformers, Precision agriculture, Transformer models, ViT, Deep learning, Plant disease detection, Plant disease classification},
}

\begin{document}
\maketitle
\begin{abstract}
Detecting plant diseases is a crucial aspect of modern agriculture, as it plays a key role in maintaining crop health and increasing overall yield. Traditional approaches, though still valuable, often rely on manual inspection or conventional machine learning techniques, both of which face limitations in scalability and accuracy. Recently, \glspl*{vit} have emerged as a promising alternative, offering advantages such as improved handling of long-range dependencies and better scalability for visual tasks. This review explores the application of \glspl*{vit} in precision agriculture, covering a range of tasks. We begin by introducing the foundational architecture of \glspl*{vit} and discussing their transition from \gls*{nlp} to \gls*{cv}. The discussion includes the concept of inductive bias in traditional models like \glspl*{cnn}, and how \glspl*{vit} mitigate these biases. We provide a comprehensive review of recent literature, focusing on key methodologies, datasets, and performance metrics. This study also includes a comparative analysis of \glspl*{cnn} and \glspl*{vit}, along with a review of hybrid models and performance enhancements. Technical challenges such as data requirements, computational demands, and model interpretability are addressed, along with potential solutions. Finally, we outline future research directions and technological advancements that could further support the integration of \glspl*{vit} in real-world agricultural settings. Our goal with this study is to offer practitioners and researchers a deeper understanding of how \glspl*{vit} are poised to transform smart and precision agriculture.
\end{abstract}

\keywords{Vision Transformers \and Precision agriculture \and Transformer models \and ViT \and Deep learning \and Plant disease detection \and Plant disease classification}

\section{Introduction}
\paragraph{}
Plant diseases pose a significant challenge to global agriculture, affecting crop yields, food security, and the economic well-being of farmers. Early and accurate detection is essential to effectively manage these diseases and minimize their impact \cite{intro-foodsecurity}. Traditionally, the diagnosis of plant diseases has depended on labor-intensive manual inspections by experts or on conventional  \gls*{ml} techniques, both of which have inherent limitations. Manual inspection is not only time-consuming but also subject to human error and inconsistency, making it impractical for large-scale agricultural operations \cite{intro1, intro-manual-limits}. Conventional \gls*{ml} methods are largely dependent on manual feature extraction, which limits their capacity to generalize across diverse environmental conditions, such as variations in lighting, background, and plant species. Moreover, these techniques are unable to effectively capture complex spatial patterns directly from raw image data \cite{intro-n, intro3}.  To address these limitations, researchers are increasingly turning to automated solutions that harness recent advances in \gls*{ml} and \gls*{cv}. 

\gls*{cnn}, a widely used traditional \gls*{dl} method, have shown some success in automating the disease detection process \cite{intro2,intro-deepsurvey}. This has created a growing demand for more robust, scalable, and intelligent systems for plant disease detection.
In response, several studies have explored the potential of \gls*{dl} to meet these challenges. Some existing surveys provide general overviews of \gls*{dl} approaches, highlighting their advantages over traditional methods and outlining the progress made in applying these techniques to plant disease detection tasks. For example, one study \cite{related-s1} reviews how \gls*{dl} and \gls*{ml} techniques have been employed for plant disease detection from image data. The study describes the transition from handcrafted features with the traditional \gls*{dl} methods to deep models like \glspl*{cnn} and segmentation models. It also discusses the implementation of object detection algorithms like YOLO and Faster R-CNN, and segmentation architectures like U-Net and Mask R-CNN, for pixel-level accuracy-based detection quality improvement.
Another work \cite{related-s2} discusses the application of \gls*{ml} and \gls*{dl} techniques in the early detection and classification of plant diseases, highlighting their importance in improving agricultural yield and reducing crop loss. It presents an extensive list of \gls*{ml} techniques such as Naïve Bayes, k-Nearest Neighbors, Decision Trees, Support Vector Machines, Random Forests, and Multi-layer Perceptrons, and explores their applications in plant disease classification using environmental and image inputs. The study also discusses preprocessing techniques and image segmentation methods, demonstrating how \gls*{ml} has enabled the automation of disease diagnosis and soil testing from models trained with crop data. The contrast between classical \gls*{ml} and \gls*{dl} methods shows that \gls*{dl} outperforms \gls*{ml} in terms of accuracy and scalability. This study emphasizes the importance of addressing current deficiencies and proposes \gls*{dl} as a promising direction for creating automatic, systematic, and scalable plant disease diagnostic systems that support both scientists and farmers in agricultural decision-making.
Another overview of studies \cite{related-s3} explores the use of \gls*{dl} techniques for plant disease detection and management in agriculture. It addresses key challenges including dataset quality, imaging sensors, model generalizability, estimation of severity, and performance comparison with human capabilities. Commonly used datasets are categorized, along with their limitations in practical scenarios. The study outlines various \gls*{dl} techniques, with a particular focus on \gls*{cnn}-based models, employed for image classification, object detection, and semantic segmentation. It highlights limitations in the ability of existing models to generalize to field conditions, primarily due to overfitting and imbalanced data. Additionally, significant research areas are highlighted, such as universal severity estimation, multi-disease detection, and model generalization across diverse conditions.
A separate study \cite{related-s4} explores the application of \gls*{dl} in plant disease recognition and classification, emphasizing its critical role in agricultural productivity and food security. It illustrates how \gls*{dl} improves upon traditional image processing techniques through automated feature extraction and enhanced objectivity and efficiency. The paper discusses foundational concepts, model performance criteria, and the evolution of \gls*{dl} in plant disease recognition. It also addresses the need for data augmentation to manage limited datasets and the use of visualization methods to improve model interpretability. The study concludes with future challenges, such as the need for more varied datasets, enhanced model robustness, and the integration of hyperspectral imaging for early detection. 

Recent advancements in  \gls*{dl} have demonstrated the potential of \glspl*{cnn} and their variants in accurately detecting and classifying plant diseases. However, despite their success, \gls*{cnn}-based models often struggle to capture global contextual information due to their inherent inductive biases and locality constraints.
The introduction of the Transformer architecture has  transformed the field of \gls*{nlp}, enabling significant advancements in machine translation, text summarization, and language modeling, and setting new performance benchmarks across a wide range of language understanding tasks \cite{intro-largelanguagemodels}. By leveraging parallel processing and self-attention mechanisms, Transformers excel at capturing long-range dependencies and contextual relationships. This breakthrough has inspired researchers to explore the application of Transformers in other domains, including \gls*{cv}. It was soon recognized that the same properties that made Transformers effective in \gls*{nlp}, namely their ability to model long-range dependencies and contextually relevant information, could also be advantageous for image analysis tasks \cite{vit2020}. Unlike \glspl*{cnn}, which rely on inductive biases such as local connectivity and spatial hierarchies, Transformers offer a more flexible architecture with fewer assumptions about input structure. This flexibility enables them to learn rich, complex patterns directly from visual data, making them a promising alternative for various  \gls*{cv} applications, including automated plant disease detection. 

The use of \glspl*{vit} in agriculture is a novel and rapidly emerging research area. To the best of our knowledge, there are a few works whose focus is on \glspl*{vit}; however, most of them mention it only briefly and do not provide a comprehensive analysis of their applications in precision agriculture.
For example, a review \cite{related-s5} categorizes \gls*{dl} techniques applied to plant disease detection, organizing the literature into three tasks: classification, detection, and segmentation of plant diseases from leaf images. The review evaluates a broad range of \gls*{dl} models, with a strong focus on \glspl*{cnn} such as ResNet~\cite{resnet}, VGG~\cite{vgg}, and EfficientNet~\cite{efficientnetb0}, alongside growing interest in \glspl*{vit}. Another study \cite{related-s6} examines the use of \gls*{dl} and \gls*{cv} in plant disease detection, systematically analyzing the transition from traditional image processing to modern \gls*{dl}-based approaches. It assesses the suitability and performance of image processing techniques like thresholding, edge detection, and region-based segmentation in real-world agriculture. The study categorizes various \gls*{dl} architectures, such as \glspl*{cnn}, \gls*{gan}, \glspl*{vit}, and Vision-Language Models, and highlights their scalability and potential for practical implementation. It covers advanced models like YOLO and Faster R-CNN for real-time object detection, and discusses the emerging relevance of \glspl*{vit} for capturing global context in image analysis. However, while \glspl*{vit} are introduced, the review lacks a deep evaluation of specific architectures or performance benchmarks in agricultural contexts.

In this work, we examine the use of \gls*{vit} in the agriculture domain with the objective of identifying how this architecture has been used and adapted for agricultural applications such as crop classification, disease detection, yield prediction, and precision agriculture. With \gls*{vit}'s success in \gls*{cv} across numerous domains, its application to transforming agricultural image analysis holds much promise. By analyzing the existing literature, we aspire to extract current trends, evaluate model performance, and identify challenges specific to agricultural environments and datasets. This initial exploration will lay the groundwork for future studies and build a sound body of knowledge within this emerging interface of \gls*{dl} and agriculture. 

In summary, the key contributions of this work are as follows:
\begin{itemize} 
    \item An overview of the Transformer and \gls*{vit} architectures and their functionality is provided.
    \item We have comprehensively addressed the concept of inductive biases and their role in \gls*{ml} models, especially \glspl*{cnn} and \glspl*{vit}.
    \item A comprehensive survey is presented on the application of \glspl*{vit} in the context of precision agriculture, systematically analyzing 44 high-impact peer-reviewed studies.
    \item We critically analyze the current challenges and unresolved issues in the field, highlight emerging research trends, formulate key open challenges, and suggest promising directions for future investigation.
\end{itemize}

The structure of this paper is organized as follows: Section \ref{background} provides an overview of the Transformer architecture, followed by a discussion on \glspl*{vit}. Section \ref{inductive bias} explores the concept of inductive biases in \gls*{ml}, with a particular focus on how these biases manifest in \glspl*{cnn} and \glspl*{vit}. In Section \ref{applications}, we review the applications of \glspl*{vit} in precision agriculture, categorizing the models into two groups: pure and hybrid models. Section \ref{findingchallenges} highlights key findings and outlines the open challenges identified in the reviewed literature. Finally, Section \ref{conclusion} presents the conclusions of the paper.

\section{Background}
\label{background}
\subsection{Transformer}
\paragraph{}
The Transformer model~\cite{attention2017} brought a groundbreaking shift to \gls*{nlp} by introducing the self-attention mechanism, which allows the model to evaluate the importance of different words in a sentence during sequence encoding and generation. This innovation overcomes the limitations of earlier architectures, such as \glspl*{rnn} and \glspl*{cnn}, which often struggled with long-range dependencies and parallelization. By processing all tokens in a sequence simultaneously, self-attention captures long-range relationships more effectively, making the Transformer architecture both efficient and highly scalable for a wide range of sequence modeling tasks. In the following subsection, we examine its main components.

\subsubsection{Encoder-Decoder structure}
\begin{figure}
    \centering
    \includegraphics[width=0.5\linewidth]{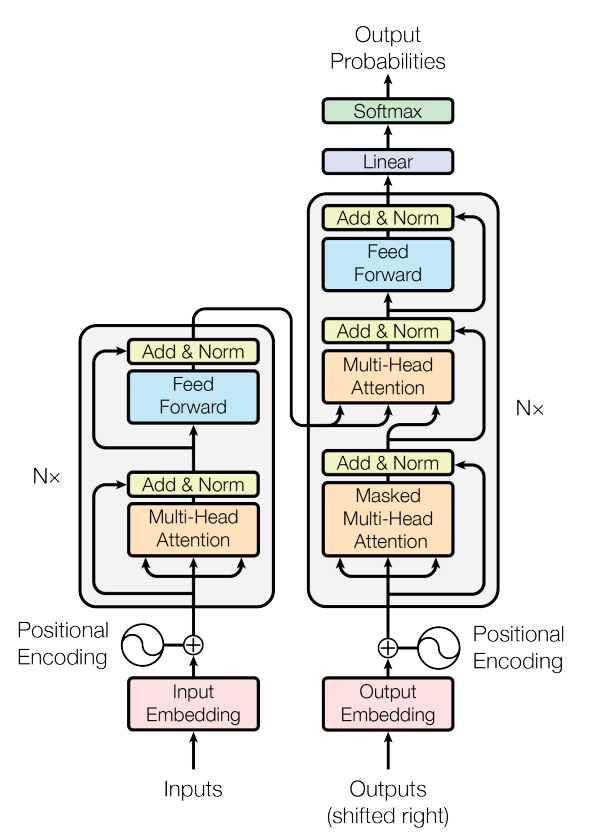}
    \caption{The Encoder-Decoder architecture of the Transformer model \cite{attention2017}. The encoder processes the input sequence to generate contextualized representations, which are then passed to the decoder to produce the output sequence. This structure enables efficient parallel processing and captures long-range dependencies through self-attention mechanisms, allowing for better modeling of relationships between distant elements in the sequence.}
    \label{fig:fig1}
\end{figure}

\paragraph{}
The Transformer architecture comprises an encoder and a decoder, each consisting of multiple layers of self-attention and Feed-Forward Neural Networks. The encoder processes the input sequence into a contextual representation, while the decoder generates the output sequence based on both the encoder’s output and previously generated tokens. This encoder–decoder structure is commonly used in tasks such as machine translation, whereas certain variants, such as GPT~\cite{gpt}, employ only the decoder for autoregressive language modeling. Figure~\ref{fig:fig1} illustrates the overall architecture of the Transformer’s encoder–decoder structure.

\vspace{\baselineskip}
Each encoder layer consists of two main sublayers:
\begin{enumerate}
    \item Self-Attention Mechanism
    \item Feed-Forward Neural Network
\end{enumerate}

Each sublayer is followed by a residual connection and layer normalization.
\vspace{\baselineskip}

Each decoder layer consists of three sublayers:
\begin{enumerate}
    \item Masked Self-Attention Mechanism
    \item Encoder-Decoder Attention Mechanism
    \item Feed-Forward Neural Network
\end{enumerate}

Similar to the encoder, each sublayer is followed by a residual connection and layer normalization.

\subsubsection{Self-Attention mechanism}
\begin{figure}[htb]
    \centering
    \includegraphics[width=0.3\linewidth]{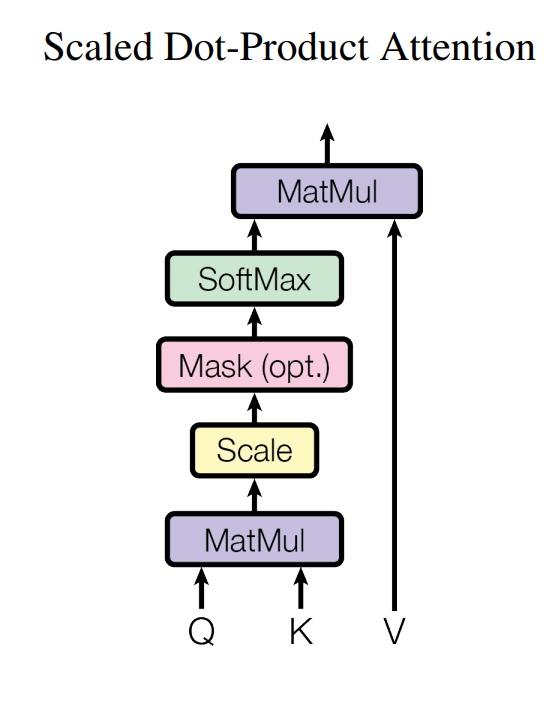}
    \caption{The Attention mechanism \cite{attention2017}. The attention computation uses queries (Q), keys (K), and values (V) to generate weighted representations, allowing the model to focus on relevant parts of the input sequence based on learned attention scores.}
    \label{fig:attention}
\end{figure}

\paragraph{}
The self-attention mechanism is the core innovation of the Transformer architecture, allowing the model to dynamically focus on different parts of the input sequence. As shown in Figure~\ref{fig:attention}, this attention is computed using the following steps:

\begin{enumerate}
    \item Input Matrices: The inputs consist of queries (Q), keys (K), and values (V), all derived from the input embeddings.
    \item Attention Scores: Calculate the dot products of the query (Q) with all keys (K). This results in a matrix of raw attention scores, which determines how much attention each query should pay to the different keys.
    \item Scaling: Scale the attention scores by the square root of the dimension of the keys, \(\sqrt{d_{k}}\), to avoid large values that could result in very small gradients during training.
    \item Softmax: Apply the softmax function to the scaled attention scores to normalize them, turning them into probabilities (attention weights) that sum to 1.
    \item Weighted Sum: Compute the weighted sum of the values (V) using the attention weights. The result is a new set of embeddings (or representations), where each one is a mixture of the original values weighted by how much attention was paid to the corresponding key.
\end{enumerate}

\subsubsection{Multi-Head Attention}
\begin{figure}[htb]
    \centering
    \includegraphics[width=0.33\linewidth]{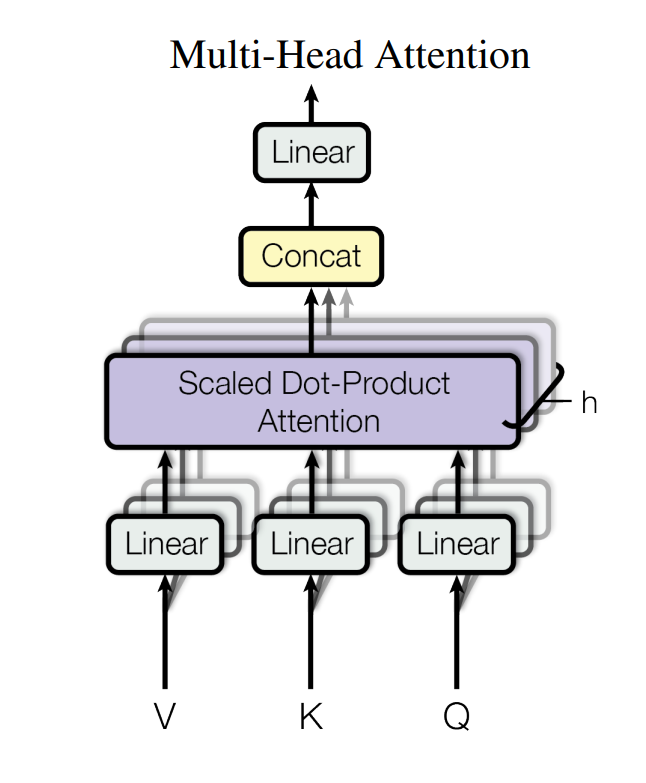}
    \caption{multi-head attention \cite{attention2017}. It runs multiple Self-Attention operations in parallel, each with different learned projections of Q, K, and V, allowing the model to capture diverse contextual relationships from different representation subspaces.}
    \label{fig:mha}
\end{figure}

\paragraph{}
To allow the model to focus on different parts of the sequence simultaneously, the Transformer uses multi-head attention, which involves multiple parallel attention layers (heads). The process can be summarized as follows:

\begin{enumerate}
    \item Linear Projections: Linearly project the queries, keys, and values \(h\) times using different learned projection matrices for each head, resulting in \(h\) distinct sets of queries, keys, and values.
    \item  Parallel Attention Heads: Apply the scaled dot-product attention mechanism independently to each set of queries, keys, and values in parallel. This allows the model to attend to different parts of the input sequence from different subspaces.
    \item  Concatenation: Concatenate the outputs of all \(h\) attention heads, forming a single vector of combined information.
    \item Final Linear Projection: Apply a final linear transformation to the concatenated output, resulting in the final multi-head attention output that will be passed to subsequent layers. 
\end{enumerate}
Figure \ref{fig:mha} provides a schematic representation of the multi-head attention mechanism.

\subsection{Vision Transformer}
\begin{figure}[htb]
    \centering
    \includegraphics[width=0.8\linewidth]{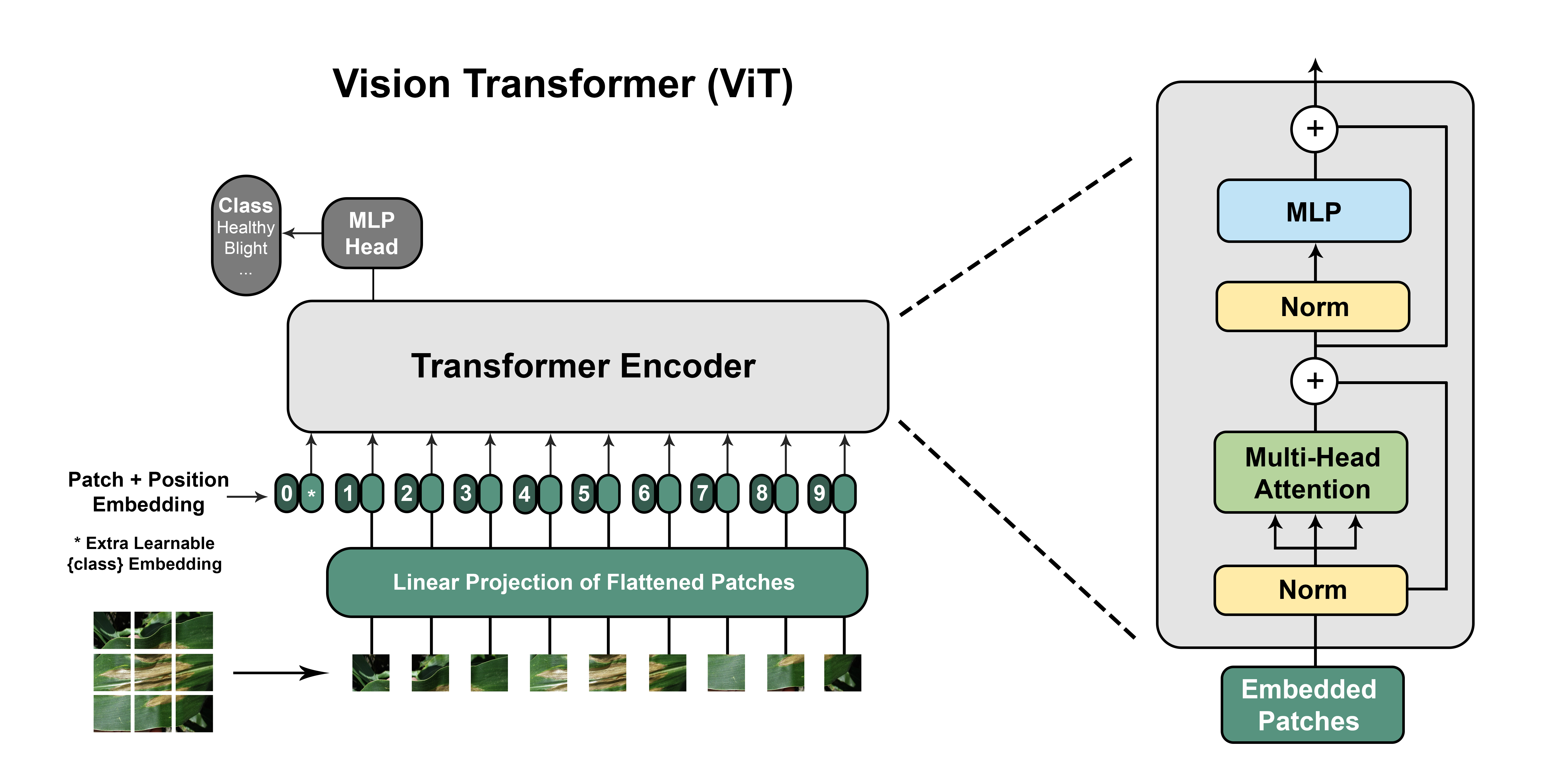}
    \caption{Vision Transformer architecture \cite{vit2020}. The image is split into fixed-size patches, which are linearly embedded, combined with positional embeddings, and fed into a Transformer encoder. The model captures global image context through Self-Attention across all patches.}
    \label{fig:vit}
\end{figure}

\paragraph{}
Inspired by the success of Transformers in \gls*{nlp}, A. Dosovitskiy et al.~\cite{vit2020} adapted the Transformer architecture for \gls*{cv} tasks, leading to the creation of \gls*{vit}. Before this breakthrough, visual attention mechanisms were typically combined with convolutional networks or used to enhance specific components, while retaining the overall \gls*{cnn} structure. Early attention-based architectures had shown promise in \gls*{cv}~\cite{squeeze}, but their implementation on hardware accelerators often required complex and specialized engineering, as these systems were primarily optimized for convolutional operations. The key insight of \gls*{vit} was to show that convolutional networks are not essential: a simple Transformer applied directly to sequences of image patches can achieve strong results in image classification. In \gls*{vit}, images are divided into fixed-size patches, each treated as a token, similar to words in NLP. These patches are linearly embedded, combined with positional embeddings, and processed through multiple Self-Attention layers. This design enables \glspl*{vit} to effectively capture long-range dependencies and contextual relationships within images, leading to performance that often surpasses traditional \glspl*{cnn} on various classification benchmarks. By leveraging the Self-Attention mechanism of Transformers, \glspl*{vit} have revolutionized the field of \gls*{cv}. The following section delves into the architecture of \glspl*{vit}, explaining their key components and operations in detail.
Figure~\ref{fig:vit} shows the structure of \gls*{vit}. The architecture of \gls*{vit} consists of the following key components: Patch Embedding, Positional Embedding, Transformer Encoder, and Classification Head.

\subsubsection{Patch Embedding}
In \glspl*{vit}, the input image is divided into fixed-size patches. Each patch is then flattened and projected into a lower-dimensional space.\newline
Given an image \(X\in \mathbb{R}^{H\times W\times C} \)
 with height \(X\), width \(W\), and \(C\) channels, it is divided into patches of size \(P\times P\) Thus, the number of patches \(N\) is \(\frac{HW}{P^{2}} \). \newline
Each patch is then linearly transformed into an embedding vector using a trainable projection matrix \(E\in \mathbb{R}^{(P^{2}\cdot C)\times D}\) where \(D\) is the dimensionality of the embedding space.
 
\begin{itemize}
    \item The embedding dimension \(D\) in a \gls*{vit}  plays a crucial role in transforming the flattened image patches into a suitable format for processing by the Transformer architecture. 
    \item An embedding is a vector representation of the input data. For image patches, it means converting the flattened pixel values into a high-dimensional vector that captures more meaningful information.
    \item Directly using the raw pixel values of image patches may not be efficient or effective for the Transformer model. Embedding helps transform these raw values into a more abstract, dense representation.
    \item The choice of the embedding dimension \(D\) in a \gls*{vit} is a crucial design decision that balances the model's representational power and computational efficiency. 
\end{itemize}

Let  \(z^{P}_{0} \in \mathbb{R}^{D}\) denote the initial embedding of the \(P^{th}\) image patch. It is obtained by flattening the patch and projecting it into a D-dimensional space using a learnable linear transformation:

\begin{equation}
    z^{P}_{0} = \text{Flatten}(x^{P}) E, \quad \text{for } P = 1 \text{ to } N
\end{equation}

where \(z^{P}_{0} \in \mathbb{R}^{D}\) is the embedding of the \(P^{th}\) patch at Transformer layer-\(0\) (before any attention layers), \(x^{P}\)  is the  \(P^{th}\) image patch before projection and \(E \in \mathbb{R}^{(P^2.C)\times D}\) is a learnable weight matrix that projects each flattened image patch into a \(D\)-dimensional embedding space.

\subsubsection{Positional Embedding}
To retain the spatial information, the positional embedding \(P\in \mathbb{R}^{N\times D}\) is added to the patch embeddings: 

\begin{equation}
    z_{0} = [x_{1}E\ ;\ x_{2}E\ ;\ \dots\ ;\ x_{N}E] + P
\end{equation}
where \(z_{0}\) is the full embedded input sequence after adding positional information at Transformer layer-\(0\) (before any attention layers), \(x_{i} \in \mathbb{R}^{P^2.C}\) is the \(i^{th}\) flattened image patch, \(E \in \mathbb{R}^{(P^2.C)\times D}\) is the patch embedding projection matrix and \(P \in \mathbb{R}^{N\times D}\) is the positional embeddings. The positional embeddings added to the patch embeddings can be either learned or fixed. 

\subsubsection{Transformer Encoder}
The Transformer encoder consists of multiple layers, each composed of a multi-head self-attention mechanism and a feed-forward neural network.\newline

\textbf{1-Multi-Head Self-Attention (MHSA):}

The self-attention mechanism computes a weighted sum of the input sequences, focusing on different parts of the sequence for each output element. Given an input sequence \(z\in \mathbb{R}^{N\times D}\)

\begin{equation}
    \text{Attention}(Q, K, V) = \text{softmax}\left( \frac{QK^{T}}{\sqrt{D}} \right) V
\end{equation}

where \(Q=zW_{Q}\), \(K=zW_{K}\), and \(V=zW_{V}\) are query, key, and value matrices,
respectively. \(W_{Q}\), \(W_{K}\) and \(W_{V}\) are learned projection matrices.\newline
For multi-head attention, multiple attention heads are used:

\begin{equation}
    \text{MHSA}(z) = [\text{head}_{1}\ ;\ \text{head}_{2}\ ;\ \dots\ ;\ \text{head}_{h}] W_{O}
\end{equation}

where \( \text{head}_{i} = \text{Attention}(Q_{i}, K_{i}, V_{i})\) and  \(W_{O} \in \mathbb{R}^{hD \times D}\) is a learned projection matrix.\newline

\textbf{2- Feed-Forward Neural Network (FFN)}

Each encoder layer also includes a feed-forward neural network, applied to each position separately and identically:

\begin{equation}
    \text{FFN}(x) = \max(0,\, x W_{1} + b_{1}) W_{2} + b_{2}
\end{equation}

where \(W_{1}\in \mathbb{R}^{D\times D_{ff}} \) and \(W_{2}\in \mathbb{R}^{D_{ff}\times D}  \) are learned weight matrices, \(b_{1}\in \mathbb{R}^{D_{ff}}\) and \(b_{2}\in \mathbb{R}^{D}\) are learned bias vectors, and \(D_{ff}\) is the dimensionality of the feed-forward layer.\newline

\textbf{3- Layer Normalization and Residual Connections}

Each sub-layer (MHSA and FFN) is followed by layer normalization and a residual connection:

\begin{equation}
    z'_{l} = \text{LayerNorm} \left( \text{MHSA}(z_{l-1}) + z_{l-1} \right)
\end{equation}

\begin{equation}
    z_{l} = \text{LayerNorm} \left( \text{FFN}(z'_{l}) + z'_{l} \right)
\end{equation}

where \(z_{l-1}\) is the output from the previous encoder layer, \(MHSA(z_{l-1})\) is the result of applying multi-head self-attention to that output, \(z^{'}_{l}\) is the input to the feed-forward network (FFN) sub-layer.\(z_{l}\) denotes the output of $l^{th}$layer.

\subsubsection{Classification Head}
The output of the Transformer encoder is a sequence of embeddings, one for each input token. For classification tasks, a special learnable classification token \(z^{0}_{0}\) is prepended to the input sequence before encoding. After passing through all \(L\) encoder layers, the corresponding output embedding \(z^{0}_{L}\) is used as a compact representation of the entire image. This embedding is then passed through a multi-layer perceptron (MLP) to produce the final class scores:

\begin{equation}
    \text{logits} = \text{MLP} \left( z^{0}_{L} \right)
\end{equation}

where the output \(\text{logits} \in \mathbb{R}^{K}\) (for \(K\) classes) is used for prediction.

\section{Inductive biases in ML and DL Models}
\label{inductive bias}
\subsection{Definition of Inductive Biases}
\paragraph{}
Inductive biases refer to a set of built-in assumptions that a \gls*{ml} model creates, which can facilitate the generalization process, especially when working with limited data. We can consider these biases as prior knowledge that the model uses to make predictions about data that it hasn't seen before. In general terms, inductive biases guide the learning algorithm to solutions that possess particular characteristics \cite{baxter2000model, goyal2022inductive}.

In supervised learning, when the data is frequently incomplete or noisy, inductive biases are essential for the model’s effectiveness. In\gls*{cv}, for example in \glspl*{cnn}, some of the inherent inductive biases include locality and translation equivariance \cite{defferrard2016convolutional,ravanbakhsh2017equivariance,finzi2020generalizing}. Locality assumes that important features, such as edges and textures, are restricted to small spatial regions within an image, while translation equivariance guarantees that these features are identified regardless of their position in the image. These biases enable CNNs to perform exceptionally well in tasks like image classification and object detection, where spatial hierarchies play a key role \cite{inductive-imagenet,inductive-yolo}.  Similarly, in \gls*{nlp}, \glspl*{rnn} and Transformers utilize sequential and contextual relationships as biases to process and understand text. The type of inductive biases has a direct impact on a model’s performance on particular tasks.

\subsection{Inductive Biases In CNNs and ViTs}
\paragraph{}
\textbf{CNNs} have demonstrated remarkable effectiveness in modeling spatial hierarchies and capturing local dependencies within data. Their architecture, which consists of convolutional layers followed by pooling layers, makes them well-suited for image recognition, object detection, and segmentation, where learning spatial relationships is essential \cite{ib-semantic, ib-pooling}. The efficiency of \glspl*{cnn} is largely attributed to the strong inductive biases natively built into their design, including locality, two-dimensional neighborhood structure, and translation equivariance \cite{vit2020}. These are uniformly applied across all layers in the network, enabling \glspl*{cnn} to learn invariant and spatially-conscious feature representations.

In particular, two dominant inductive biases, locality and weight sharing, serve as architectural constraints that significantly reduce the number of learnable parameters and enhance the model’s generalization capabilities \cite{ib-convit}. Locality is a design principle whereby each neuron in a convolutional layer connects only to a small, spatially localized subset of the input, known as the receptive field. This localized computation enables \glspl*{cnn} to effectively extract low-level features such as edges, corners, and textures. As data passes through deeper layers, local features are hierarchically combined to detect more abstract and global representations \cite{ib-deeplearning}. Weight sharing, meanwhile, ensures the same filters (kernels) are employed at different spatial locations, which makes the model equivariant to translation. As a result, inductively learned patterns in one region of an image are recognizable wherever they are, thereby making the model stronger and more efficient, significantly reducing the number of trainable parameters, enabling it to generalize well, and improving computational efficiency \cite{ib-equivariant}. Together, these inductive biases not only reduce the model’s complexity but also add prior knowledge about the structure of visual data, which accelerates training and improves performance, especially in scenarios with limited labeled data. Although CNNs' intrinsic inductive biases can be very effective when data is insufficient, they can be limiting when there is sufficient data. In this context, it is necessary to select the most effective inductive biases rather than employing all available ones \cite{ib-convit}. The benefit of inductive biases tends to diminish with very large datasets, which implies that transfer learning scenarios, where only a limited number of examples are available for the new distribution, provide a valuable context for assessing the effectiveness and implementation of these biases \cite{goyal2022inductive}. It is due of these characteristics that a randomly initialized \gls*{cnn} can perform well in object localization tasks without any learning \cite{ib-randomcnn}. Despite all these advantages, the locality characteristic of \glspl*{cnn} causes them to degrade in their ability to capture long-range dependencies \cite{ib-convit}.

\textbf{ViTs} have demonstrated remarkable effectiveness in modeling spatial hierarchies and capturing global dependencies within data. Unlike CNNs, which are specifically designed to operate with local data by spatially bounded filters, \glspl*{vit} depend on self-attention mechanisms where every token (or image patch) can attend to any other token in the image regardless of their spatial positions \cite{vit2020}. In the \gls*{vit} architecture, while \gls*{mlp} layers function locally and maintain a degree of translational equivariance, the self-attention layers are inherently global, enabling \glspl*{vit} to integrate information across the entire spatial domain at every layer. Rather than relying on the strong inductive biases embedded within CNNs, such as locality, translation equivariance, and weight sharing, ViTs replace these constraints with global processing achieved through multi-head self-attention. This design allows \glspl*{vit} to learn flexible and context-dependent representations by dynamically attending to different parts of the image. As a result, \glspl*{vit} are capable of incorporating broader contextual cues at earlier stages in the network, which can lead to superior performance in tasks that benefit from understanding long-range relationships between image regions. On the other hand, CNNs must build up such spatial relationships incrementally through repeated layers of local convolutions, making the learning process more architecture-biased than data-driven.

However, this lack of built-in inductive biases also presents some challenges. The absence of explicit mechanisms for capturing local details can make \glspl*{vit} less effective at modeling fine-grained patterns, especially in the early layers \cite{ib-dovisioncnn}. Additionally, \glspl*{vit} exhibit high sensitivity to the choice of hyperparameters, such as the optimizer, learning rate schedule, and network depth, which makes their training more delicate and often requires large-scale datasets and computational resources \cite{ib-earlyconvoltion}. Surprisingly, despite these architectural differences, recent results show that \glspl*{vit} can implicitly learn some of the inductive biases that are commonly found in \glspl*{cnn}. For instance, early self-attention heads in \glspl*{vit} prefer to attend to local regions, mimicking the effect of convolutional kernels \cite{ib-earlyconvoltion}. This observation shows that with sufficient training data and good optimization, \glspl*{vit} are capable of learning some of the beneficial priors that \glspl*{cnn} have by nature. While \glspl*{vit} offer a more flexible and globally conscious architecture for visual representation learning, they lack the strong, human-crafted inductive biases that cause CNNs to excel so thoroughly in data-scarce regimes. This trade-off between flexibility and architectural bias underscores the importance of understanding when and how to use \glspl*{vit} effectively, particularly with respect to the scale and nature of the data involved.

\subsection{Strategies to Mitigate the Lack of Inductive Biases in ViTs}\paragraph{}
\glspl*{vit} lack the explicit inductive biases that are inherent in \glspl*{cnn}, such as locality and weight sharing, which can make it challenging for them to learn effectively, especially with limited data. One approach to mitigate this lack of inductive biases involves injecting convolutional inductive biases into \glspl*{vit}. This can be done through convolutional patch embedding, which modifies how image data is turned into input tokens for the Transformer. Specifically, it replaces or enhances the original flat patch embedding with convolutional layers to capture local spatial patterns more effectively. Alternatively, a convolutional stem can be added at the beginning of the network to create hybrid models, helping the model learn local features more effectively.

Another strategy involves using knowledge distillation from a teacher model that possesses strong inductive biases. Distilling the knowledge in a neural network involves training a smaller neural network, known as the student or distilled model, to replicate the behavior of a larger, more complex model or ensemble of models, often referred to as the teacher or cumbersome model. The main goal is to transfer the broader understanding and generalization capability learned by the larger model into the smaller one \cite{knowledgesitilling}. In this approach, one or more pre-trained \gls*{cnn} models are often used as the teacher to transfer their inductive biases to the \gls*{vit} student model \cite{ib-mitigate-distill}. This method has been shown to improve the performance of \glspl*{vit}, particularly on smaller datasets.

Using hierarchical structures is another way to mitigate the lack of inductive biases by introducing a multi-stage architecture that processes visual information similarly to \glspl*{cnn}. Instead of attending globally from the start, these models apply self-attention within local windows and gradually merge or downsample tokens across layers, creating a spatial hierarchy \cite{swinvit}. This setup embeds locality and spatial coherence, which are key inductive biases of CNNs, allowing the model to initially capture low-level features locally and subsequently build more abstract, global representations \cite{ib-mitigate-pvt}.
 Therefore, the most common approaches used to mitigate the lack of inductive biases in \glspl*{vit} can be summarized as follows:

\begin{itemize}
    \item \textbf{Convolutional Patch Embedding:} It involves replacing simple patch embedding with convolutional layers to preserve local structures in the image, or employing recursive token aggregation, which functions similarly to convolutions in capturing local spatial relationships. This strategy is utilized in models such as Tokens-to-Token ViT (T2T-ViT) \cite{ib-mitigate-T2T}.
    
    \item \textbf{Hybrid Models:} One way to mitigate the lack of inductive biases in \glspl*{vit} is to use a \gls*{cnn} or convolutional backbone to extract low-level features and then feed this output into a \gls*{vit}. This approach combines the spatial inductive biases of \glspl*{cnn} with the global modeling capabilities of \glspl*{vit}. Introducing a few convolutional inductive biases in the early stages of \gls*{vit} processing can strike a balance between inductive bias and the learning capacity of Transformer blocks, while also influencing the optimization behavior of \glspl*{vit} \cite{ib-earlyconvoltion}.
    
    \item \textbf{Knowledge Distillation:} Distilling knowledge from \glspl*{cnn} into \glspl*{vit} during training is another possible solution, as exemplified by DEiT \cite{deit}, which uses a \gls*{cnn} as the teacher and a \gls*{vit} as the student to improve performance through teacher–student learning during the training process.
    
    \item \textbf{Hierarchical Structure:} Designing \glspl*{vit} with hierarchical architectures to process multi-scale features is what \gls*{swint} does, which results in improved efficiency and local feature awareness \cite{swinvit}.
\end{itemize}

\section{Applications of ViTs in Precision Agriculture}
\label{applications}
\paragraph{}
\glspl*{vit} have been increasingly applied to plant disease detection, leveraging their powerful feature extraction and attention mechanisms. Various studies have explored different adaptations of \glspl*{vit} to improve performance in this domain. These adaptations can be broadly categorized into two types: those that use \glspl*{vit} and its variants in their pure form or with minor modifications, which we categorize as \textbf{Pure Models}, and those that implement significant changes to the architecture and combine it with convolutional characteristics, which we categorize as \textbf{Hybrid Models}.

\subsection{Pure Models}
\paragraph{}
In this section, we review studies that employ pure \gls*{vit} models or their variants with minor modifications. These modifications include fine-tuning or other adjustments that maintain the overall \gls*{vit} structure but aim to improve the model's performance. In Table \ref{tab:tablepure}, we provide a summary of such models. The main goals of these studies are to implement task-specific models through fine-tuning on specific datasets, to reduce the number of model parameters to decrease resource consumption, and to create lightweight models for real-world deployment.

A simple use case of \gls*{vit} was shown in a study that proposed a \gls*{vit}-based approach for the automatic classification of strawberry diseases using transfer learning and fine-tuning techniques \cite{P-n2}. The authors enhanced the \gls*{vit} model by adding new layers such as ReLU activation, batch normalization, dropout, and a softmax classifier. Comparative evaluations against conventional \gls*{cnn} models like VGG16, VGG19~\cite{vgg}, ResNet50V2~\cite{resnetv2}, and MobileNet~\cite{mobilenet} showed that the proposed \gls*{vit} model, especially when combined with data augmentation, significantly outperformed other methods. In their best setting, the model achieved an accuracy and F1-score of 92.7\% on the test set. The research highlighted the strengths of Transformer-based architectures in precision agriculture, especially in complex visual tasks such as disease recognition from strawberry leaves, fruits, and flowers. The study also evaluated the proposed method on the PlantVillage dataset to assess generalizability, where it attained 98.9\% accuracy, outperforming both the original \gls*{vit} and previous state-of-the-art methods like Mask R-CNN~\cite{mask-rcnn} and MCLCNN. The authors concluded that the \gls*{vit} model’s ability to capture global image features made it a powerful alternative to \glspl*{cnn} for agricultural disease detection tasks.

In another study, GreenViT \cite{p2-2023}, a novel \gls*{vit}-based model, was proposed to improve plant disease detection by addressing the limitations of traditional \glspl*{cnn}. Unlike \glspl*{cnn}, which may lose spatial information due to pooling layers, GreenViT segmented input images into smaller patches to effectively capture essential features. The model was evaluated using three datasets: PlantVillage (PV), Data Repository of Leaf Images (DRLI), and a new Plant Composite (PC) dataset created by merging the first two datasets. PlantVillage consisted of 54,305 images spanning 38 classes (26 infected, 12 healthy). DRLI included 4,502 images of 12 plant species. GreenViT achieved impressive accuracy rates of 100\% on PV, 98\% on DRLI, and 99\% on PC, outperforming state-of-the-art models such as VGG16, MobileNetV1, and \gls*{vit}-Base. The model’s efficiency was highlighted by its reduced parameter count (21.65 million compared to 86 million in \gls*{vit}-Base), making it computationally lightweight and suitable for edge devices.

\begin{figure}[htb]
    \centering
    \includegraphics[width=0.8\linewidth]{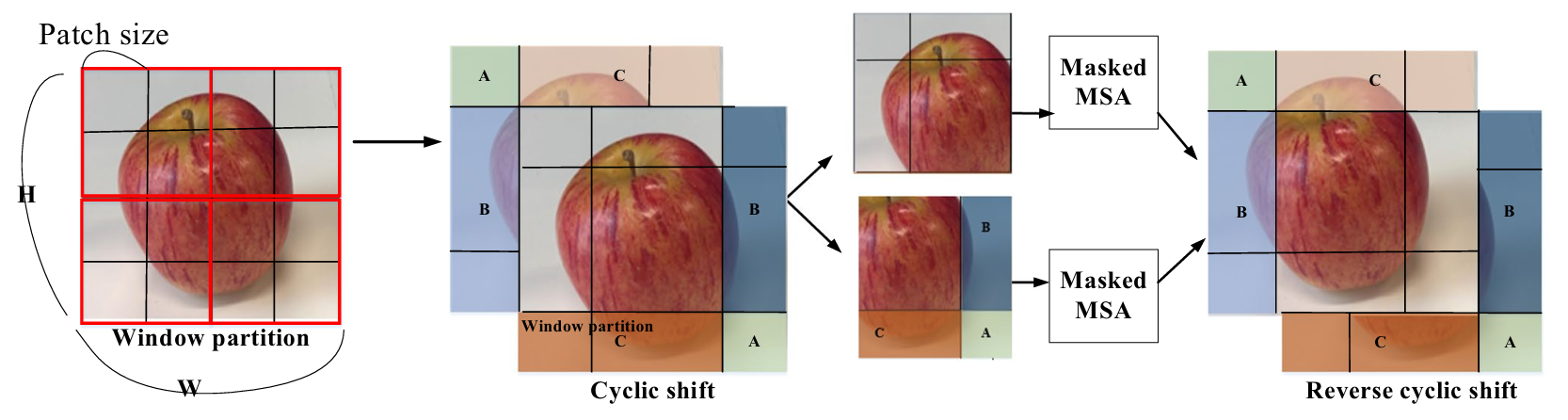}
    \caption{The Shifted window mechanism in Swin Transformer based on MSA \cite{p44-2021}. The red box indicates a window where self-attention is applied, while black boxes represent individual image patches. The shaded region B illustrates the area that is masked out during shifted window self-attention to preserve the locality of each window.}
    \label{fig:pureswint}
\end{figure}

As discussed in Section~\ref{inductive bias}, using a hierarchical architecture was one of the possible solutions to mitigate the lack of inductive biases in \glspl*{vit}. \gls*{swint} is a type of \gls*{vit} designed to address challenges associated with traditional \glspl*{vit}, such as the high computational cost of processing high-resolution images and the need for a hierarchical structure to effectively capture features at multiple scales \cite{swinvit}. Several studies employed the \gls*{swint} model in precision agriculture applications. An application of \gls*{swint} was introduced in a study~\cite{p44-2021} that examined Transformer models for accurate fruit ripeness classification. The authors addressed the problem of performing fruit selection and pickup effectively in the absence of professional labor. They created their own fruit datasets, including apples and pears, to train, test, and compare various Transformer models, like \gls*{vit}, \gls*{swint}, and \gls*{mlp}-based models. The \gls*{mlp} model performed well but was prone to overfitting when it had higher capacity. The moving window-based self-attention in \gls*{swint}, as shown in Figure~\ref{fig:pureswint}, facilitated better feature extraction of visual objects. The experiments demonstrated that \gls*{swint} yielded the best results, achieving a precision of 87.43\% for fruit object detection. This work also combined the Transformer module and YOLO~\cite{yolo} module to effectively classify the ripeness stages of apples and pears.

\begin{figure}[htb]
    \centering
    \includegraphics[width=0.8\linewidth]{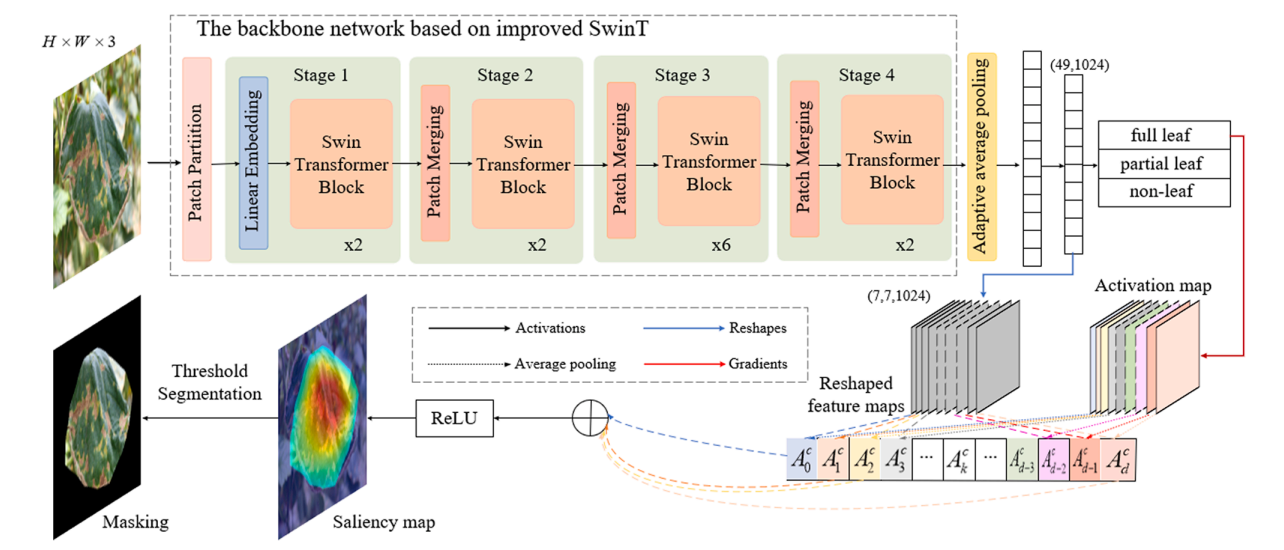}
    \caption{Overview of the proposed leaf extraction module \cite{p47-2022}. It includes an improved \gls*{swint}-based backbone for leaf region determination, Grad-CAM for generating a saliency map, and threshold segmentation for extracting the leaf region from complex backgrounds.}
    \label{fig:pure2}
\end{figure}

\begin{figure}[htb]
    \centering
    \includegraphics[width=0.8\linewidth]{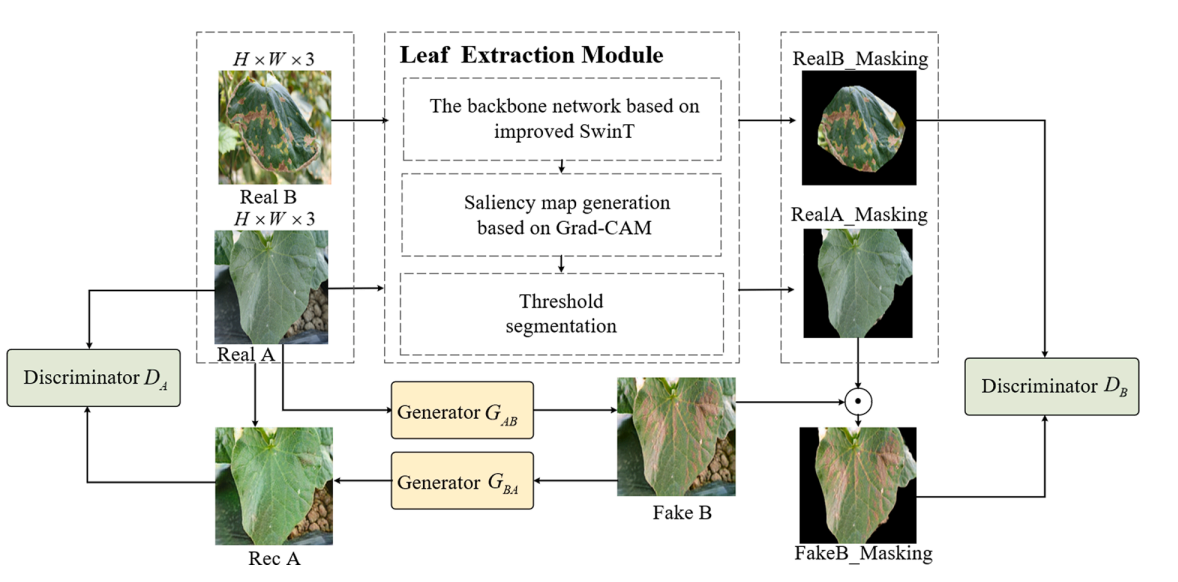}
    \caption{The framework of the proposed STA-\gls*{gan} \cite{p47-2022}. The generator transforms healthy images (A) into diseased images (B) using a series of convolutional, residual, and transposed convolutional layers. Leaf region masks guide both the generator and discriminator to focus on leaf areas, enhancing disease translation accuracy.}
    \label{fig:pure3}
\end{figure}

An improved framework of the \gls*{swint} architecture~\cite{p47-2022} was proposed to identify cucumber leaf diseases, addressing challenges such as complex backgrounds and limited sample sizes. The authors developed a \gls*{swint}-based and attention-guided \gls*{gan} (STA-\gls*{gan}) model capable of generating diseased images without altering background integrity. The leaf extraction module (Figure~\ref{fig:pure2}), which included the proposed backbone network and Grad-CAM~\cite{gradcam}, was incorporated into the \gls*{gan} to create STA-\gls*{gan} (Figure~\ref{fig:pure3}). They employed a modified \gls*{swint} model, which used an improved feature extraction technique of step-wise small patch embeddings and Coordinate Attention (CA), without adding any extra parameters. Their results showed that STA-\gls*{gan} and the improved \gls*{swint} had the potential to minimize dependency on large manually annotated datasets, while at the same time improving disease detection.

There are some studies whose main contribution is the use of new self-built or custom datasets. For instance, a customized \gls*{vit} model~\cite{p32-2024} was developed to identify and categorize Java Plum leaf diseases. The dataset used in that experiment was self-collected and included six classes: Bacterial Spot, Brown Blight, Powdery Mildew, Sooty Mold, Dry, and Healthy leaves. The study showcased the capabilities of \gls*{vit} for improving plant disease detection technologies with respect to the example of Java Plum crops. The model developed in that work surpassed the performance of traditional \gls*{cnn}-based models and achieved higher detection accuracy. The use of \glspl*{vit} for plant disease detection using multispectral images captured in natural environments was explored in a recent study~\cite{p12-2023}. That work aimed to assess the performance of \glspl*{vit} for plant disease recognition applied to multispectral imagery acquired under natural conditions. The authors created a novel dataset named the “New Multispectral Dataset”, which contained pictures of five agricultural plants: avocados, tomatoes, lime, passionfruit, and gooseberries. This database included pictures taken under both visible and near-infrared (NIR) capture modes, thus offering more information for disease analysis than RGB-only capture. Four \glspl*{vit} models, namely \glspl*{vit}-S16, \glspl*{vit}-B16, \glspl*{vit}-L16, and \glspl*{vit}-B32, were trained from scratch and subsequently fine-tuned using ImageNet-21k weights. The research examined the roles of model size, patch size, and transfer learning on the models’ ability to classify diseases. The findings indicated that the \gls*{vit} model based on the B16 architecture and augmented with transfer learning provided the best results, with training and testing accuracies of 93.71\% and 90.02\%, respectively.

In addition to utilizing a self-built dataset, a novel residual-distilled Transformer architecture~\cite{p38-2023} was proposed for the identification of rice leaf diseases. The authors argued that existing works on rice leaf disease identification were unsatisfactory in terms of either accuracy or model interpretability. To address these issues, they introduced a method that combined a \gls*{vit} and a distilled Transformer in a residual block for feature extraction, followed by a \gls*{mlp} for prediction. The researchers collected their own dataset on rice leaf disease between 2019 and 2020. This dataset, which they referred to as the rice leaf disease dataset, consisted of pictures of four rice leaf diseases: bacterial blight, brown spots, blast, and tungro. The proposed method achieved an 89\% F1-score and 92\% top-1 accuracy on the rice leaf disease dataset, outperforming other state-of-the-art models.

In another work, LEViT~\cite{p39-2024} was proposed as a model based on \gls*{vit} for identifying and classifying leaf diseases in agriculture, with an emphasis on enhancing interpretability through Explainable AI techniques. Integrating Grad-CAM provided a visual explanation of the model’s predictions in the form of important regions in the images of the leaves. Their model achieved high performance, with a training accuracy of 95.22\%, validation accuracy of 96.19\%, and test accuracy of 92.33\%. Notably, Grad-CAM visualizations enhanced the trustworthiness and usability of the model by elucidating its decision-making processes. Their findings showed that LEViT’s advanced structure helped to improve leaf disease recognition and diagnosis, which was essential in precision farming. The integration of interpretability enabled LEViT to address real-world issues of early disease detection and management.

As we mentioned, one of the approaches that can be taken is to create efficient models by reducing parameters or employing transfer learning strategies. An innovative transfer learning strategy designed for plant disease recognition in scenarios with limited data was proposed in a recent study~\cite{p27-2022}. Pre-training was performed using the PlantCLEF2022 dataset, which contained 2,885,052 images across 80,000 classes~\cite{PlantCELF2022}. The authors employed a \gls*{vit} model that was initially self-trained on ImageNet and subsequently trained on the PlantCLEF2022 dataset. Several benchmark models, such as \glspl*{cnn} and other \gls*{vit}-based systems, were used to validate the proposed method. The results indicated that the model outperformed the previous models in all dataset configurations, including few-shot learning scenarios. For instance, it achieved remarkable results even when trained with only 20 images per class, and it also performed well in plant growth stage prediction and weed recognition. Moreover, the dual transfer learning approach improved accuracy while also demonstrating faster convergence, showcasing the strength of the method across various dataset types. This study underscored the effectiveness of large-scale, plant-related datasets combined with advanced models like \gls*{vit} for transfer learning. It emphasized the necessity of appropriate source datasets, self-supervised pre-training, and fine-tuning processes for plant disease recognition.

In another work~\cite{p7-2023}, the \gls*{swint} network was used to improve the identification of weeds in agriculture. To address challenges posed by limited training data at scale, a two-stage transfer learning strategy was employed. First, the authors pre-trained the \gls*{swint} model on the ImageNet dataset and then fine-tuned it on the Plant Seedlings dataset as well as on a private dataset named MWFI (Maize/Weed Field Image). This strategy leveraged pre-trained network parameters to specialize in feature extraction for weed recognition. The proposed method achieved a recognition accuracy of 99.18\%, outperforming other \gls*{cnn}-based architectures such as EfficientNetV2~\cite{efficientnetv2}.

LeafViT, a \gls*{vit}-based model tailored for detecting leaf diseases in plants, specifically targeted fine-grained features critical for classification~\cite{p35-2022}. Experiments underscored the robustness of LeafViT across various configurations, validating its computational efficiency relative to models such as VGG16. The authors highlighted its potential for deployment on edge devices and its adaptability for object detection and segmentation tasks. Another work~\cite{p3-2024} introduced a novel application of \gls*{vit} models for detecting diseases in tomato plants, leveraging smartphone-based technology. A smartphone application developed as part of that study integrated the \gls*{vit} model for real-time, field-level disease detection. By deploying the model on a smartphone application via TensorFlow Lite, that work provided a practical tool for field use, enabling farmers with real-time insights to enhance crop health and yield.

In another study, IEM-ViT was proposed as an innovative approach to detect tea leaf diseases~\cite{p29-2023}. It incorporated an information entropy weighting (IEW) method to prioritize regions within images that held significant information, enhancing the feature extraction process. This model used a masked autoencoder (MAE) with an asymmetric encoder-decoder architecture, effectively masking 75\% of the input image to filter out redundant and irrelevant background data. This method enabled proper model development and facilitated accurate disease recognition even when low-quality images were used. The findings indicated that the IEM-ViT model surpassed previously developed algorithms. Therefore, IEM-ViT was considered suitable for fast and large-scale tea disease recognition and was proposed as deployable for recognizing diseases in other crops, contributing to enhanced agricultural management and sustainability.

FormerLeaf~\cite{p17-2023}, a Transformer-based cassava leaf disease classification model, was introduced in a recent study. The model consisted of 12 Transformer encoder layers with optimized self-attention mechanisms, which helped reduce resource usage while maintaining high classification accuracy. To enhance FormerLeaf, two optimization techniques were proposed: the Least Important Attention Pruning (LeIAP) algorithm and Sparse Matrix-Matrix Multiplication (SPMM). LeIAP identified and removed less critical attention heads, resulting in a 28\% reduction in model size and a 15\% increase in inference speed without significant accuracy loss. SPMM reduced computational complexity from O(n2)O(n2) to O(n2/p)O(n2/p), improving training efficiency by 10\% and lowering resource consumption. These improvements made FormerLeaf suitable for real-world applications in precision agriculture, where computational efficiency was critical. The authors also mentioned challenges, including dataset imbalance and variable image quality due to diverse lighting conditions. To address these, they suggested using advanced augmentation techniques, such as \glspl*{gan}~\cite{gan}, to improve model performance.

{\scriptsize
\begin{longtable}{p{0.04\textwidth} p{0.18\textwidth} p{0.08\textwidth} p{0.08\textwidth} p{0.08\textwidth} p{0.08\textwidth} p{0.28\textwidth}}
\caption{Summary of pure models. These models employ the pure form of ViT, aiming to enhance its performance or fine-tune it. (Reported metrics are directly extracted from the original publications. In certain studies, per-class metrics are provided; metrics that are not reported are indicated with a "-".)} \label{tab:tablepure}  \\
\toprule
\textbf{Paper} & \textbf{Dataset(s)} & \textbf{Accuracy} & \textbf{Precision} & \textbf{Recall} & \textbf{F1} & \textbf{Main Contribution} \\
\midrule
\endfirsthead

\toprule
\textbf{Paper} & \textbf{Dataset(s}) & \textbf{Accuracy} & \textbf{Precision} & \textbf{Recall} & \textbf{F1} & \textbf{Main Contribution} \\
\midrule
\endhead

\midrule
\multicolumn{7}{r}{\textit{Continued on next page}} \\
\endfoot

\bottomrule
\endlastfoot
\cite{P-n2} & PlantVillage and Strawberry Disease Detection dataset & 92.7\% & - & - & 0.927 & Fine-tuning a ViT model with additional dense and normalization layers for precise strawberry disease classification \\
        \midrule
		\cite{p44-2021} & Self-built & - & 0.8743 & - & - & Adopt \gls*{swint} for accurate ripeness classification of various fruit types \\
        \midrule
		\cite{p47-2022} & Self-built & 98.97\% & - & 0.9873 & - & Improving patch partition of \gls*{swint} and using \gls*{swint}-based Attention-guided \gls*{gan} \\
        \midrule
		\cite{p32-2024} & Self-built & 97.51\% & Per-class & Per-class & Per-class & Introducing customised ViT model for effective classification of Java Plum leaf diseases \\
        \midrule
		\cite{p38-2023} & Self-built & 92\% & 0.88 & 0.91 & 0.89 & Residual-distilled Transformer architecture for rice leaf disease identification \\
        \midrule
		\cite{p39-2024} & New Plant Diseases Dataset & 92.33\% & - & - & - & ViT-based model incorporating Explainable AI (XAI) features \\
        \midrule
		\cite{p27-2022} & PlantCLEF2022 & Varies & - & - & - & Novel transfer learning strategy for versatile plant disease recognition \\
        \midrule
		\cite{p7-2023} & Plant Seedling and a self-built dataset & 99.18\% & 0.9933 & 0.9911 & 0.9922 & Fine-grained recognition using \gls*{swint} and two-stage transfer learning \\
        \midrule
		\cite{p35-2022} & Plant Pathology 2020 & 97.15\% & Per-class & Per-class & Per-class & ViT-based approach to identify and amplify subtle discriminative regions \\
        \midrule
		\cite{p3-2024} & PlantVillage & 90.99\% & Per-class & Per-class & Per-class & Smartphone-based plant disease detection \\
        \midrule
		\cite{p2-2023} & \makecell[tl]{PlantVillage \\ DRLI \\ PlantComposite} & \makecell[tl]{100\% \\ 98\% \\ 99\%} & \makecell[tl]{- \\ - \\ -} & \makecell[tl]{- \\ - \\ -} & \makecell[tl]{- \\ - \\ -} & Reducing number of learning parameters \\
        \midrule
		\cite{p29-2023} & Tea leaves disease classification (Kaggle) & 93.78\% & 0.9367 & 0.9380 & 0.9364 & Asymmetric encoder-decoder architecture for masked autoencoder with selective patch encoding and pixel reconstruction \\
        \midrule
		\cite{p17-2023} & Cassava Leaf Disease Dataset (Kaggle) & 98.52\% & - & - & 0.9682 & Proposing LeIAP algorithm for selecting key attention heads in Transformer \\
\end{longtable}}

\subsection{Hybrid Models}
\paragraph{}
In this section, we review models that use ViT with specific adjustments, particularly those that combine ViT with certain \gls*{cnn} characteristics. A summary of the papers is shown in Table~\ref{tab:tablehybrid}.

In a comparative study~\cite{p5-2022}, the authors evaluated the performance of standalone and hybrid models and proposed a lightweight \gls*{dl} approach utilizing \gls*{vit} for automated plant disease classification. They compared \gls*{vit}-based models, \glspl*{cnn}, and hybrid architectures that combined both techniques. These models were evaluated on three different datasets: the Wheat Rust Classification Dataset (WRCD), the Rice Leaf Disease Dataset (RLDD), and the PlantVillage dataset. A \gls*{vit}-based model achieved a 100\% F1-score on WRCD with a 200×200 resolution, outperforming EfficientNetB0~\cite{efficientnetb0} in computational efficiency. For RLDD, hybrid models balanced accuracy and computational cost, with the \gls*{vit}-based model achieving the highest average precision. On the larger PlantVillage dataset, the \gls*{vit}-only model reached an F1-score of 98.77\%, surpassing \gls*{cnn}-based architectures. Models that combined attention blocks and \glspl*{cnn} as hybrid models offered a balance between accuracy and speed. They were comparatively faster than pure \gls*{vit} models and more accurate than \gls*{cnn}-only models. For instance, a hybrid model achieved high accuracy in the PlantVillage experiments, indicating the potential of integrating attention mechanisms into \glspl*{cnn} for real-world applications. It is worth mentioning that this study highlighted the need for a practical, lightweight, real-time implementation for agricultural purposes. Multi-label classification with object localization for detailed disease identification was identified as the focus of future research.

EfficientRMT-Net~\cite{p48-2023} is a novel hybrid model that integrated ResNet-50 and \gls*{vit} in order to classify potato plant leaf diseases effectively. EfficientRMT-Net employs depth-wise convolution for computational efficiency and utilizes a stage-block architecture to enhance scalability and feature sensitivity. This paper emphasizes that the ability of EfficientRMT-Net to process high-dimensional data efficiently while maintaining high accuracy positions it as a valuable tool for smart farming. Another paper~\cite{p46-2023} proposed SEViT model combining Squeeze-and-Excitation~\cite{squeeze} ResNet101 and \gls*{vit} to tackle large-scale and fine-grained plant disease classification challenges. The preprocessing network employs Squeeze-and-Excitation modules in ResNet101 to enhance channel-specific feature representation, while the \gls*{vit} component leverages its global attention mechanism for feature extraction and classification. SEViT improves upon the limitations of \gls*{cnn}-based methods, which often struggle to distinguish diseases in visually similar crops. Although comparisons with popular models such as MobileNetV3~\cite{mobilenetv3}, EfficientNet, and VGG16 highlight the efficiency of SEViT in handling visually similar diseases across different crops, SEViT faces challenges including high computational requirements due to the depth of the preprocessing network and its reliance on pretrained weights to optimize the performance of \gls*{vit}. 

An enhanced \gls*{vit} model~\cite{p19-2024} based on the MaxViT~ \cite{ maxvit} architecture was introduced for identifying diseases in maize leaves. By adapting the MaxViT structure with the addition of Squeeze-and-Excitation (SE) blocks and implementing Global Response Normalization (GRN) in the \gls*{mlp} layers, the proposed model achieved significant improvements in both accuracy and inference speed. The researchers created a comprehensive maize dataset by merging the PlantVillage, PlantDoc, and CD\&S datasets, forming the largest publicly available maize disease dataset. This combined dataset, featuring four disease classes, was split into training, validation, and testing sets to rigorously evaluate the generalization capabilities of \gls*{dl} models. The model was benchmarked against 28 \gls*{cnn} and 36 \gls*{vit} architectures. It achieved a record-breaking accuracy of 99.24\% on the test set, outperforming existing methods in both accuracy and inference speed, making it particularly well-suited for real-world agricultural applications.

A Spatial Convolutional Self-Attention-based Transformer (SCSA-Transformer) \cite{p41-2023} for strawberry disease identification was proposed to address challenges such as complex image backgrounds and dataset imbalances. The architecture incorporated hierarchical feature mapping through convolutional modules and multi-head self-attention to capture dependencies across diverse image regions. Experimental results demonstrated that the proposed method achieved an accuracy of 99.10\% and underscored the SCSA-Transformer's capability for precise disease recognition, even in complex scenarios, with faster convergence and reduced computational requirements compared to alternative methods.

In another study, a novel Convolutional \gls*{swint} (CST) model was proposed to address challenges in plant disease detection \cite{p33-2022}. By integrating convolutional layers into the \gls*{swint} architecture, the CST model enhanced feature extraction from noisy and complex agricultural images. The study evaluated the model on multiple datasets: the Cucumber Plant Diseases Dataset, Banana Leaf Disease Images, Potato Disease Leaf Dataset, and a tomato subset of PlantVillage. Notably, CST maintained an accuracy of 0.795 even with 30\% salt noise, demonstrating its robustness. In addition to its strong classification performance, CST showed notable resilience to noise, outperforming baseline models such as ResNet50 and LeViT-192 under salt noise conditions. While CST outperformed existing models in noisy data accuracy, its computational complexity, which exceeded 48 million parameters, was noted as a limitation.

In a recent work~\cite{P-n4}, an automated system was proposed for identifying two prevalent mango leaf diseases, anthracnose and powdery mildew, commonly found in the Andhra Pradesh region of India. The authors employed Compact Convolutional Transformer (CCT) models, specifically CCT-7/8×2 and CCT-7/4×2, to classify diseased leaves using \gls*{dl}-based \gls*{vit} architectures. These models were selected for their efficiency in handling small datasets with reduced computational requirements. The experimental evaluation was conducted using a self-built mango leaf image dataset collected from over 50 trees. The dataset comprised 574 images labeled across four classes: healthy, dead, anthracnose, and powdery mildew. The study confirmed that compact Transformers like CCT were highly effective for disease detection in resource-limited agricultural contexts. The authors planned to extend their work by incorporating additional disease categories and expanding the dataset to improve the generalizability of the proposed model. 

A novel approach using \glspl*{vit} was introduced to address challenges in paddy leaf disease identification \cite{p4-2023}. The model incorporated a multi-scale contextual feature extraction mechanism to capture both local and global representations of lesions on paddy leaves. This was complemented by a weakly supervised Paddy Lesion Localization (PLL) unit that prioritized significant lesion areas to enhance classification accuracy. The model's performance was evaluated on the Paddy Doctor dataset \cite{paddydoctor-dataset}. The proposed model outperformed several state-of-the-art models, including DenseNet169~\cite{densenet}, ResNet50, and \gls*{swint}, achieving an accuracy of 98.74\%, an F1-score of 98.18\%, and an AUC of 99.49\%. These results demonstrated significant improvements over baseline models, highlighting the efficacy of lesion-focused feature extraction and context-aware learning in accurately classifying complex diseases. The authors emphasized that integrating lesion localization with multi-scale \gls*{vit} structures enabled robust performance across diverse visual challenges. The model's enhanced ability to identify and prioritize discriminative lesion areas provided more reliable disease detection, supporting its potential for real-world applications in precision agriculture.

A Hybrid Pooled Multihead Attention (HPMA) model~\cite{p16-2024} was proposed for agricultural pest classification, enhancing both local and global feature extraction in images. HPMA integrated hybrid pooling and \gls*{vit} techniques to address the limitations of conventional attention mechanisms and \gls*{cnn} models. It utilized a novel attention mechanism that combined max and min pooling to improve feature weighting and discriminative power. The model was tested on a newly collected dataset of 10 pest classes, achieving a testing accuracy of 98\%. Its effectiveness was further validated on two benchmark datasets: a medium dataset (MD) with 40 classes and a large dataset (LD) with 102 classes. For the MD, HPMA achieved a testing accuracy of 98.02\%, while for the LD, it reached 95.98\%. These results emphasized the model’s adaptability to varying dataset sizes and complexities. Comparisons with state-of-the-art \gls*{cnn} and Transformer models highlighted HPMA's efficiency, offering competitive accuracy with relatively lower computational requirements.

An innovative approach for accurate tomato disease identification~\cite{p31-2023} was presented through the development of the NanoSegmenter model, which integrated Transformer structures with lightweight techniques such as the inverse bottleneck, quantization, and sparse attention mechanisms. This method aimed to balance high precision with computational efficiency, addressing challenges in traditional models such as detailed instance segmentation and efficient deployment on mobile devices. The use of lightweight processing enabled smartphone integration, which was vital for practical field use by farmers, streamlining disease detection and supporting timely agricultural interventions.

The CRFormer~\cite{p26-2023} model was introduced to address challenges in segmenting grape leaf diseases from complex natural backgrounds. The model employed a unique cross-resolution mechanism to retain high-resolution (HR) and low-resolution (LR) feature maps in parallel, enhancing contextual understanding. It featured a Large-Kernel Mining (LKM) attention mechanism for adaptive spatial and channel encoding, and a Multi-Path Feed-Forward Network (MPFFN) for multi-scale representation. The model used a lightweight Hamburger decoder for effective multi-resolution data fusion. The results highlighted CRFormer’s capability to handle the complexities of real-world agricultural segmentation tasks.

EfficientNet Convolutional Group-Wise Transformer (EGWT) \cite{p21-2024} is a novel architecture for crop disease detection. It combined the convolutional capabilities of EfficientNet with the hierarchical and lightweight group-wise Transformer mechanism. This architecture addressed challenges in computational efficiency and parameter constraints while maintaining high accuracy. EGWT extracted local features via EfficientNet convolution and processed them through grouped Transformer modules, reducing redundancy and emphasizing relevant spatial relationships. The EGWT architecture was validated using three benchmark datasets: PlantVillage, Cassava, and Tomato Leaves. Visualization of intermediate layers confirmed its ability to focus accurately on disease-affected regions. Despite challenges in recognizing complex Cassava leaf patterns and imbalanced datasets, EGWT demonstrated superior performance across metrics. Its lightweight design made it an excellent choice for deployment in resource-constrained environments, potentially transforming automated crop disease diagnosis in agriculture.

Another work proposed an innovative approach to enhance plant disease diagnosis through synthetic data augmentation. The research introduced LeafyGAN~\cite{p18-2024}, a \gls*{gan} framework comprising a pix2pix \gls*{gan} for segmentation and a Cycle\gls*{gan} for disease pattern generation. By generating synthetic images to address dataset imbalances, Leafy\gls*{gan} ensured robust augmentation, producing high-quality representations of diseased leaves. A MobileViT classification model \cite{mobilevit}, selected for its lightweight architecture and computational efficiency, was trained on augmented datasets, achieving accuracies of 99.92\% on the PlantVillage dataset and 75.72\% on the PlantDoc dataset. These results highlighted the efficacy of Leafy\gls*{gan} in supporting models tailored for low-resource environments. Leafy\gls*{gan} demonstrated notable advancements over existing augmentation methods, such as the LFLSeg module and traditional \glspl*{gan}, by preserving the background during disease pattern translation and producing visually coherent images. The integration of MobileViT, with its blend of \gls*{vit} and convolutional operations, underscored the balance between performance and efficiency. Compared to resource-heavy architectures, MobileViT effectively handled the augmented datasets, demonstrating scalability for real-world applications. This framework marked a significant step in addressing data scarcity and computational limitations, paving the way for practical deployment in agriculture for disease diagnosis and prevention.

One idea for creating a hybrid model is using ensemble models. In a recent study~\cite{p45-2024}, researchers introduced the MDSCIRNet architecture, a novel \gls*{dl} model combining Depthwise Separable Convolution (DSC) and \gls*{vit}-based multi-head attention mechanisms for detecting potato leaf diseases. In addition to the standalone model, hybrid approaches integrating classical machine learning methods (e.g., SVM, Random Forest, Logistic Regression, and AdaBoost) with the MDSCIRNet model were proposed. These combinations yielded competitive results, particularly in cases leveraging ensemble learning techniques. In another study~\cite{p42-2024}, the authors introduced an ensemble learning approach with hard and soft voting strategies that integrated three \gls*{cnn} models (MobileNetV3, DenseNet201, ResNext50~\cite{resnext}) and two Transformer models (\gls*{vit} and \gls*{swint}) to classify leaf diseases . Experimental results showed that ensemble learning improved classification accuracy, with \gls*{vit} achieving superior performance as a standalone model. The authors also employed Grad-CAM visualization to highlight the regions of input images most influential for classification decisions, demonstrating that the models effectively localized disease areas. Their results indicated that ensemble learning combining \gls*{cnn} and Transformer models provided a powerful approach for effective classification of plant diseases, which was useful in practical agriculture. A novel ensemble model~\cite{p22-2024}, Residual Swin Transformer Networks (RST-Nets), was proposed to address the challenges of plant disease recognition, including noise, varying image intensities, and the subtle differences between healthy and diseased plants. The model integrated residual convolutional blocks with \gls*{swint}, leveraging the latter's hierarchical architecture for scalable complexity and global context awareness. Residual connections enhanced the feature extraction process by retaining crucial information from earlier layers. The model achieved outstanding results, surpassing state-of-the-art models such as ResNet and GoogleNet~\cite{googlenet}, but its computational overhead limited its suitability for resource-constrained IoT and edge devices.

\begin{figure}
    \centering
    \includegraphics[width=0.8\linewidth]{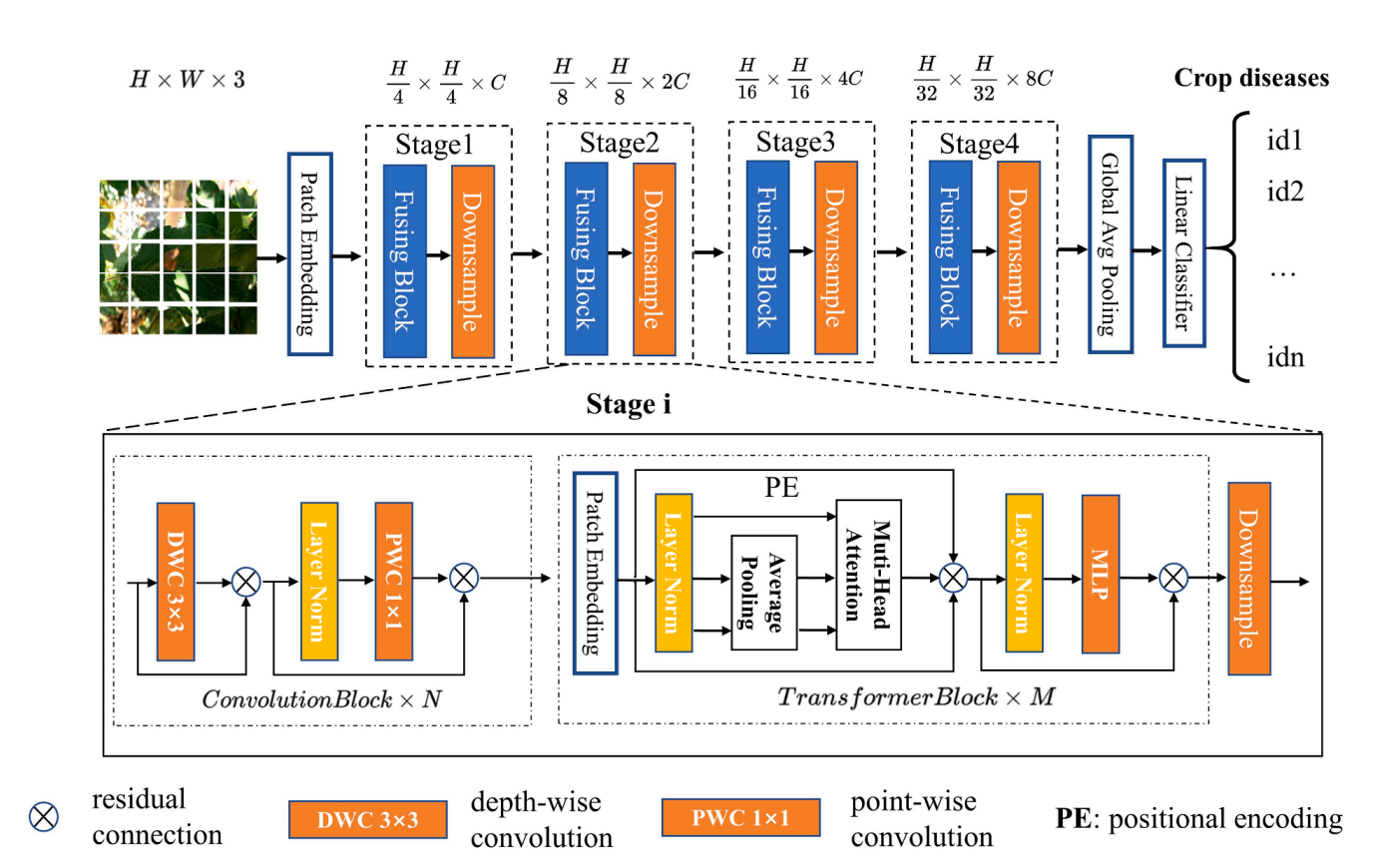}
    \caption{Overview of the ConvViT architecture \cite{p9-2022}. The top row illustrates the overall hierarchical structure of ConvViT, while the bottom row presents a detailed view of Stage 2. Each stage is composed of N convolutional blocks followed by M Transformer blocks.}
    \label{fig:hybrid1}
\end{figure}

Another way to create a hybrid model is to build a multi-stage model that combines \gls*{cnn} and \gls*{vit} by dividing the network into sequential stages. This hierarchical approach reduces computational costs while preserving essential information, enabling robust performance in complex environments. As shown in Figure~\ref{fig:hybrid1}, ConvViT~\cite{p9-2022}, which combined \gls*{vit} and \glspl*{cnn}, was developed to identify kiwifruit diseases in complex natural environments. The authors created a custom dataset of 25,168 images capturing six types of kiwifruit leaf diseases and validated the model’s performance on this dataset, as well as on the publicly available PlantVillage and AIChallenger2018 datasets. To enhance feature extraction, the model employed overlapping patch embeddings and alternating convolutional and Transformer layers, ensuring both global and local feature representation. On the kiwifruit dataset, ConvViT achieved a top identification accuracy of 98.78\%, surpassing benchmarks such as ResNet, \gls*{vit}, and ResMLP~\cite{resmlp}, while maintaining a lightweight design with reduced parameters and FLOPs. Key improvements included overlapping patch embedding to preserve local image continuity and reduce computational complexity, as well as a multi-stage design to balance feature extraction efficiency and computational cost. The Transformer’s global attention complemented \gls*{cnn}’s local feature extraction, making ConvViT highly effective in complex environments with variable lighting, backgrounds, and noise. The model’s innovative design and lightweight adaptability made it a valuable backbone for broader identification tasks in agriculture, potentially aiding real-world crop disease management efforts.

A separate study~ \cite{p14-2022} presented a hybrid model combining \gls*{cnn} with Transformer encoders to improve the accuracy and generalization of crop disease diagnosis. The authors emphasized handling complex backgrounds and disease similarities, challenges that traditional \glspl*{cnn} struggled to address. Using the PlantVillage dataset, the model achieved a high recognition accuracy of 99.62\%. For real-world applications, datasets with complex backgrounds were also used: Dataset1 (apple, cassava, and cotton diseases) and Dataset2 (apple scab, cassava brown streak, and cotton boll blight), where the proposed model attained balanced accuracies of 96.58\% and 95.97\%, respectively, demonstrating robust generalization capabilities. The key to the model's success was the integration of a Transformer encoder to capture global features, which compensated for the limitation of \gls*{cnn} in extracting global context. Furthermore, a novel hybrid loss function, combining cross-entropy with Centerloss, was introduced to optimize feature separability and reduce intra-class variance. In another paper~\cite{p24-2024} researchers integrated lightweight \glspl*{cnn} and Transformer-based frameworks, specifically the BEiT model \cite{BeiT}. This fusion leveraged the feature extraction strengths of \glspl*{cnn} and the contextual learning capabilities of Transformers to detect rice leaf diseases effectively. The model processed images by dividing them into visual tokens and applying a two-stage mechanism: reconstruction of visual elements and masked image modeling. The reconstruction stage extracted meaningful patterns, while the masked image modeling stage predicted hidden sections of the image, enhancing its understanding. To improve interpretability, the Local Interpretable Model-Agnostic Explanations (LIME) technique was employed, coupled with segmentation using Simple Linear Iterative Clustering (SLIC), which highlighted the critical regions contributing to predictions. The explainable nature of the model, supported by attention mapping and segmentation, provided transparency in decision-making, fostering trust among end-users in agricultural diagnostics. This architecture was particularly suited for deployment on resource-constrained devices, enabling real-time application in smart agriculture.

There are also studies that use transfer learning in addition to hybrid models. In a recent research, the authors proposed a hybrid model, TLMViT~ \cite{p50-2023}, for plant disease classification. They incorporated data augmentation techniques, which addressed class imbalance and mitigated overfitting, along with a two-phase feature extraction process: initial extraction using pre-trained \glspl*{cnn} models (e.g., VGG16, VGG19, ResNet50, AlexNet, and Inception V3 \cite{alexnet}), and deep feature extraction using \gls*{vit}. Final classification was conducted using an \gls*{mlp} classifier. The model achieved a validation accuracy of 98.81\% on the PlantVillage dataset  and 99.86\% on the Wheat Rust dataset. These results represented a 1.11\% and 1.099\% improvement in accuracy compared to transfer learning models without \gls*{vit}. The findings indicated the effectiveness of combining transfer learning and \gls*{vit} for deep feature extraction. The authors also highlighted that TLMViT effectively leveraged pre-trained models to reduce dimensionality and computational complexity. Another paper~\cite{p23-2023} proposed the FOTCA model, which combined the strengths of Transformers and \glspl*{cnn} for plant leaf disease detection. By integrating an Adaptive Fourier Neural Operator (AFNO) with traditional convolutional down-sampling, FOTCA effectively captured both global and local features. Key innovations in FOTCA included the use of learnable Fourier features for positional encoding, which enhanced feature representation by mapping images to the frequency domain. This approach improved robustness to variations such as rotation and scaling, addressing limitations of \gls*{vit} on small and medium-sized datasets. Transfer learning on a pre-trained ImageNet model also optimized training efficiency, enabling quick convergence.

\begin{figure}
    \centering
    \includegraphics[width=0.8\linewidth]{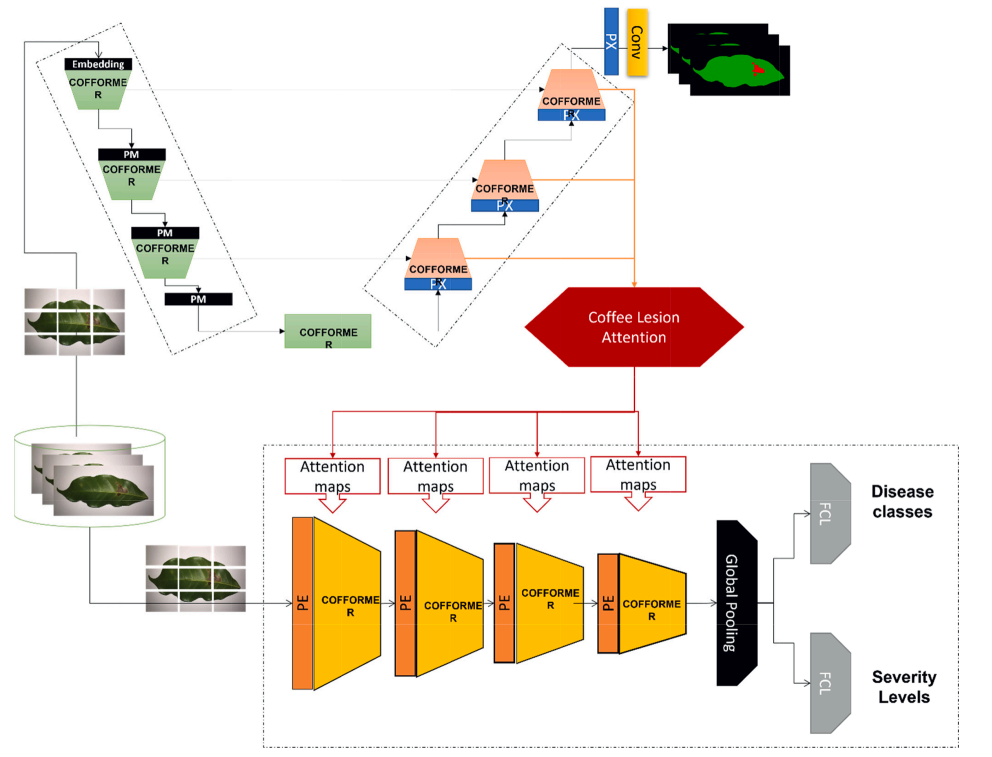}
    \caption{The architecture of COFFORMER. It consists of three main components: a segmentation subnetwork for detecting biotic stress lesions, a dual-head classification subnetwork for identifying disease type and severity, and a CLA block that enhances feature representation using guidance from the segmentation decoder \cite{p40-2024}.}
    \label{fig:hybrid3}
\end{figure}

Some studies utilize multi-track models, particularly dual-track (or dual-path) architectures, that integrate two parallel networks, such as a \gls*{vit} and a \gls*{cnn}, to leverage the strengths of both. Figure~\ref{fig:hybrid3} shows COFFORMER~\cite{p40-2024}, a novel dual-path visual Transformer designed for efficient and interpretable diagnosis of coffee leaf diseases, which integrates lesion segmentation and disease classification into a unified framework . The segmentation path employs a U-shaped architecture built on COFFORMER blocks, which use multiscale convolutional pooling for token mixing, while the classification path incorporates a Coffee Lesion Attention (CLA) module to focus on critical lesion regions. The framework evaluates both the type and severity of diseases, enabling real-time and explainable disease diagnosis.
The dataset used for evaluation consists of 1,685 images of Arabica coffee leaves captured under controlled conditions and annotated for five disease types and severity levels. Experimental results demonstrate COFFORMER's superior performance compared to state-of-the-art models. For segmentation, it achieved a Dice Similarity Coefficient (DSC) of 98.14\% and a mean Intersection over Union (mIoU) of 97.98\%, outperforming baselines such as Swin-Unet~\cite{swin-unet} and PSPNet~\cite{pspnet}. In classification tasks, COFFORMER achieved an accuracy of 99.1\% and an F1-score of 99.32\% for disease type detection, alongside 99.12\% accuracy for severity estimation. These results demonstrate its robustness in identifying diverse lesions and disease severity across real-world coffee leaf images.

For classification of citrus diseases, a novel dual-track deep fusion network~ \cite{p37-2023} was proposed that integrates a \gls*{cnn}-based Group Shuffle Depthwise Feature Pyramid (GSDFP) and a \gls*{swint}. In this architecture, the GSDFP branch extracts local features using convolutional layers and a Feature Pyramid Network (FPN) for multiscale feature fusion, while the \gls*{swint} branch captures global features through hierarchical feature maps and self-attention mechanisms. The outputs from these branches are fused and passed through a Shuffle Attention (SA) module to enhance contextual relationships between features. This dual-track design effectively combines local and global feature extraction, enabling precise classification. The proposed network achieved state-of-the-art results, outperforming existing models. It obtained a classification accuracy of 98.19\%, surpassing benchmarks such as DenseNet-121 (92\%) and AlexNet (94.3\%).  For classification of apple leaf diseases, a dual-branch model, named DBCoST~\cite{p28-2024}, integrates \gls*{cnn} and \gls*{swint}. This model includes a Feature Fusion Module (FFM) that combines information from both branches to improve accuracy. To enhance feature integration, the paper also introduces a channel attention mechanism that adjusts the importance of different feature channels. The model aims to overcome challenges such as noise from the natural environment, including overlapping branches and fruits in the images. In comparative experiments with state-of-the-art methods, including EfficientNetV2S and MobileNetV3L, DBCoST demonstrated superior performance in recognizing apple leaf diseases, particularly in challenging natural environments. In a recent paper, a hybrid \gls*{dl} model~\cite{P-n1} was introduced to detect crop diseases in complex field environments. Recognizing the limitations of conventional \gls*{cnn}-based models in handling diverse real-world scenarios, the authors proposed a dual-branch network that fused \glspl*{cnn} for local feature extraction and Transformers for global context modeling. The architecture incorporated a four-layer pyramid structure and a novel attention mechanism, using multi-head self-attention enhanced by depthwise separable convolutions and downsampling to reduce computational overhead. This design ensured that critical disease-related areas on leaves were emphasized while mitigating the influence of noisy backgrounds. To validate their model, the authors curated a custom dataset of 45,547 images depicting twelve categories of healthy and diseased crops, captured under real-world field conditions and sourced from both competitive and open datasets. Additionally, DCTN was evaluated on the publicly available CD\&S dataset, which included maize leaf diseases photographed at Purdue University's agricultural research facility. DCTN achieved state-of-the-art accuracy rates of 93.01\% on their own dataset and 99.69\% on the CD\&S dataset, outperforming popular \gls*{cnn}- and Transformer-based models such as ResNet50, EfficientNetB5, and DeiT-small. These results highlighted DCTN’s robustness and generalization capability in realistic agricultural settings, offering a promising direction for practical plant disease diagnostics. The Triple-Branch SwinT Classification (TSTC)~\cite{p34-2023} network has been proposed for the simultaneous and separate classification of plant diseases and their severity. The proposed model uses a multitask feature extraction module with a triple-branch structure, integrating \gls*{swint} for feature extraction and compact bilinear pooling (CBP) for feature fusion. It also employs a deep supervision module to enhance feature discrimination across both hidden and output layers, improving the model's accuracy and generalization. Unlike single-task models, TSTC’s multitask approach avoids species dependency and improves flexibility, addressing challenges like imbalanced datasets through robust feature fusion and supervision strategies. A critical contribution of this study lies in its end-to-end multitask framework, which leverages CBP to enhance feature fusion and deep supervision to improve training efficiency. The findings suggest that TSTC is highly effective in agricultural applications, providing accurate predictions that can inform targeted disease management strategies.

\begin{figure}
    \centering
    \includegraphics[width=0.8\linewidth]{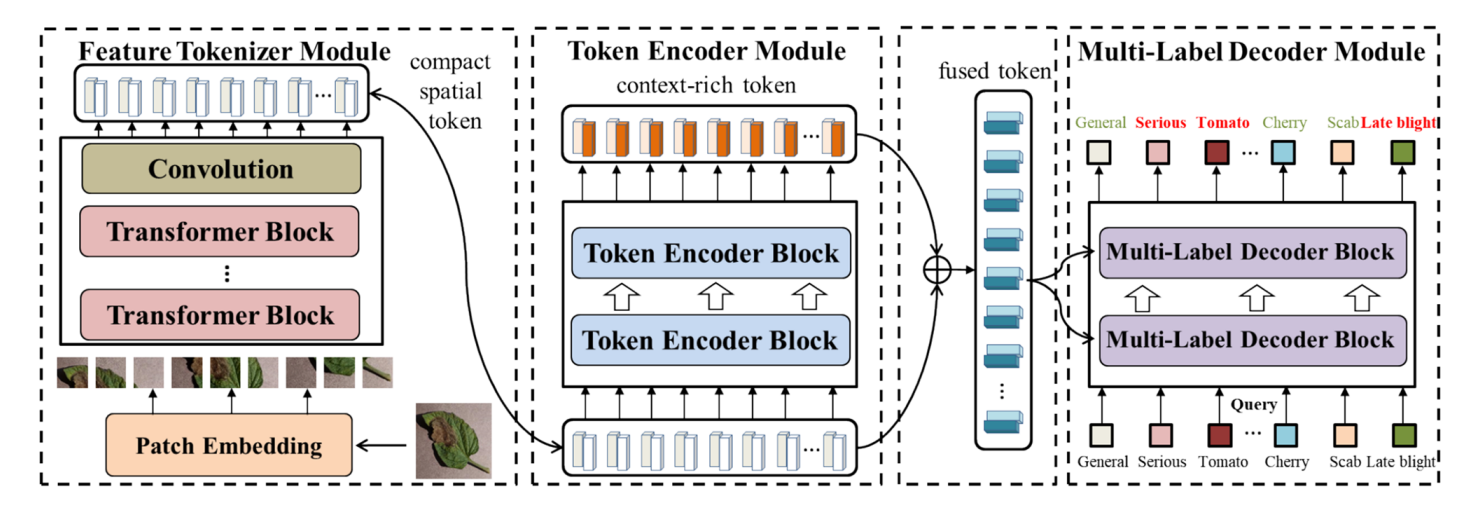}
    \caption{The architecture of LDI-NET \cite{p25-2024}. It includes a feature tokenizer module that leverages both \gls*{cnn} and Transformer strengths, a token encoder module for modeling relationships among plant type, disease, and severity, and a multi-label decoder for selectively extracting features to enable accurate multi-label identification.}
    \label{fig:hybrid2}
\end{figure}

There are still other hybrid models that introduce additional novelties. A novel multi-label model, LDI-NET, shown in Figure \ref{fig:hybrid2}, was proposed for the simultaneous identification of plant type, leaf disease, and severity \cite{p25-2024}. The model stood out due to its single-branch architecture, which avoided complex network designs and excessive categorization. LDI-NET was built around three main modules: the feature tokenizer, token encoder, and multi-label decoder. The feature tokenizer integrated the strengths of both \gls*{cnn} and Transformers, combining \gls*{cnn}’s ability to capture local details with the Transformer’s proficiency in extracting long-range global features. This module tokenized image data into compact spatial features, enhancing both local and global context awareness. The token encoder module played a crucial role in enriching the extracted tokens by facilitating information exchange among them through multi-head self-attention and \gls*{mlp} structures. This design allowed LDI-NET to capture complex relationships among plant type, disease, and severity features. The multi-label decoder module, which incorporated a residual structure, processed these context-rich tokens. It integrated shallow- and deep-level features through adaptive feature embeddings and cross-attention mechanisms to efficiently output multi-label identification results. The results underscored the model’s ability to handle multi-label identification tasks effectively, demonstrating its potential for enhancing plant disease detection in practical agricultural applications.

The Spatial Convolutional Self-Attention-based Transformer (SCSA-Transformer)~\cite{p20-2023} has been proposed to enhance strawberry disease identification under complex backgrounds. The research addresses challenges like class imbalance, limited large-scale datasets, and background complexity in agricultural disease detection. The SCSA-Transformer leverages convolutional layers to encode spatial features alongside a Transformer module for global feature extraction, improving efficiency and precision compared to existing models. The proposed model reduced the number of parameters by nearly half compared to the \gls*{swint}, facilitating lighter deployment and faster training. These enhancements underscore the method's applicability to real-world agricultural scenarios, particularly for identifying diseases with diverse and complex visual backgrounds.

A real-time plant disease identification framework called CondConViT~\cite{P-n3} has been proposed, which integrates a \gls*{vit} with a Conditional Convolutional Neural Network (CondConv) and is further enhanced by a novel attention mechanism known as Conditional Attention with Statistical Squeeze-and-Excitation (CASSE). To improve generalization and robustness, the authors introduced a data augmentation technique based on a modified Cycle\gls*{gan}, termed C3\gls*{gan}, which synthesized realistic in-field images from healthy plant samples. The model was designed to be lightweight, with only 0.95 million parameters, and was deployable on edge devices for drone-based surveillance. CondConViT was evaluated against seven state-of-the-art models and demonstrated superior interpretability and classification performance across diverse environmental conditions using both Grad-CAM and LIME techniques. They used six datasets, including five public datasets: PlantVillage \cite{plantvillage-dataset}, Embrapa \cite{Embrapa-dataset}, Plantpathology \cite{plantpathology-dataset}, Maize and Rice \cite{Maize-and-Rice-dataset}, and a newly created in-field drone image dataset named IIITDMJ-Maize \cite{IIITDMJ-Maize-dataset}, which comprised 416 labeled images of maize plants captured under various weather conditions. C3\gls*{gan} was employed to augment the IIITDMJ-Maize dataset by generating 300 synthetic images for three maize diseases: common rust, northern leaf blight, and gray leaf spot. The CondConViT model achieved the highest or near-highest performance across all datasets in terms of accuracy, F1-score, and AUC. Furthermore, it exhibited strong generalization on unseen raw drone images and outperformed heavier models, making it suitable for real-time and resource-constrained agricultural applications. In another study~\cite{pnn-5}, a parallel fusion of \gls*{cnn} and \gls*{vit} modules was proposed to capture both local and global features. Unlike sequential \gls*{cnn}-\gls*{vit} hybrids, this approach enables independent feature extraction followed by fusion using MobileNetV2~\cite{mobilenetv2} and Squeeze-and-Excitation (SE) blocks, thereby improving efficiency and interpretability. ConViTX is ultra-lightweight, comprising only 704,882 trainable parameters and requiring 0.647 GFLOPs, making it ideal for deployment on resource-constrained IoT/edge devices. Explainability was achieved using Grad-CAM and LIME, and the model’s feature discrimination capability was further validated through t-SNE visualizations.
The model was rigorously evaluated on five datasets: PlantVillage, Embrapa, PlantDoc, a custom in-field maize dataset (IIITDM-Maize), and a combined multi-species dataset called PlantCOMB. On PlantVillage, ConViTX achieved an accuracy of 99.63\%, an F1-score of 99.64\%, and an AUC of 99.96\%, outperforming nine state-of-the-art (SOTA) models. On Embrapa, it achieved an F1-score of 95.12\% and an AUC of 99.74\%, and on PlantDoc, it retained the top spot in most metrics. On the custom maize dataset, it achieved 93.02\% accuracy, matching the best competitors while showing a 2\% higher AUC. On raw drone images, it reached 61.42\% accuracy, surpassing most SOTA models in generalizability. Beyond static evaluation, ConViTX was deployed on a Jetson Nano-equipped drone and tested on real-time video frames captured from maize farms. It processed frames at approximately 29 FPS and demonstrated competitive results, notably identifying northern leaf blight with 58\% accuracy and healthy plants with 73\% accuracy. Though challenges remain with subtle diseases like common rust, the model achieved a better balance between speed, accuracy, and interpretability compared to other lightweight and heavyweight models. Overall, ConViTX proves to be an efficient, interpretable, and deployable solution, opening a promising avenue for hybrid \gls*{cnn}-\gls*{vit} architectures in real-world smart agriculture systems. RepAgrViT~\cite{pnn-6}, a novel lightweight hybrid architecture, was proposed to integrate \gls*{cnn} and \gls*{vit} principles for plant disease recognition in Internet of Things (IoT) edge environments. Unlike traditional \gls*{cnn}-\gls*{vit} hybrids, RepAgrViT employs a unique dual-stream design consisting of a token mixer and a channel mixer arranged in series to simultaneously process local and global features. A core innovation is the Bilinear Attention Transformation (BATransform) module, which captures long-range dependencies between diseased and healthy leaf regions. Additionally, the model integrates structural reparameterization techniques to reduce inference complexity while preserving accuracy. At just 9.6M parameters, the model demonstrates strong computational efficiency while retaining competitive performance. To optimize hyperparameters efficiently, the authors introduce Lite-AVPSO, a novel Particle Swarm Optimization variant that incorporates adaptive weighted delayed velocity and neighborhood-based local search mechanisms. This approach significantly enhances the hyperparameter search process over alternatives like TPE, GA, and Hyperband. Through an extensive optimization process spanning over 200 trials, the model achieved 98.54\% average accuracy across three plant disease datasets: the Coffee Leaf Dataset , the Plant Pathology Dataset , and the Rice Leaf Disease Dataset. On these datasets, RepAgrViT achieved 98.03\%, 96.99\%, and 98.78\% accuracy respectively.  Overall, this work contributes a highly efficient and interpretable AI system tailored for precision agriculture in resource-constrained settings.

{\scriptsize
\begin{longtable}{p{0.04\textwidth} p{0.18\textwidth} p{0.08\textwidth} p{0.08\textwidth} p{0.08\textwidth} p{0.08\textwidth} p{0.28\textwidth}}
\caption{Summary of hybrid models. These models are designed to leverage the complementary strengths of ViT and CNN architectures. (Reported metrics are directly extracted from the original publications. In certain studies, per-class or scenario-specific metrics are provided; metrics that are not reported are indicated with a "-".)} \label{tab:tablehybrid} \\
\toprule
\textbf{Paper} & \textbf{Dataset(s)} & \textbf{Accuracy} & \textbf{Precision} & \textbf{Recall} & \textbf{F1} & \textbf{Main Contribution} \\
\midrule
\endfirsthead

\toprule
\textbf{Paper} & \textbf{Dataset(s}) & \textbf{Accuracy} & \textbf{Precision} & \textbf{Recall} & \textbf{F1} & \textbf{Main Contribution} \\
\midrule
\endhead

\midrule
\multicolumn{7}{r}{\textit{Continued on next page}} \\
\endfoot

\bottomrule
\endlastfoot

\cite{p5-2022} & Wheat Rust Classification Dataset (WRCD), the Rice Leaf Disease Dataset (RLDD), and PlantVillage & Scenario-specific & Scenario-specific & Scenario-specific & Scenario-specific & Proposing lightweight \gls*{dl} approach based on \gls*{vit}, \gls*{cnn}, and hybrid models for real-time automated plant disease classification \\
\midrule
\cite{p48-2023} & PlantVillage & 99.24\% & Scenario-specific & Scenario-specific & Scenario-specific & Proposing hybrid model of ResNet-50 and ViT for classifying potato plant leaf diseases \\
\midrule
\cite{p46-2023} & Gathered from search engines & 88.34\% & 0.8833 & 0.8761 & 0.8750 & Combining Squeeze-and-Excitation ResNet101 and ViT for large-scale and fine-grained disease classification \\
\midrule
\cite{p19-2024} & Combination of PlantVillage, PlantDoc, and CD\&S & 99.24\% & 0.9915 & 0.9937 & 0.9926 & Adapting MaxViT structure with SE blocks and implementing Global Response Normalization in \gls*{mlp} layers for 4-class maize data \\
\midrule
\cite{p41-2023} & Self-built & 99.10\% & 0.9777 & 0.9847 & 0.9775 & Proposing SCSA-Transformer to solve strawberry disease recognition under complex backgrounds \\
\midrule
\cite{p33-2022} & Cucumber, Banana, Potato, Tomato datasets & Different in model variants & - & - & - & Proposing Convolutional Swin Transformer (CST) model for robust and accurate plant disease detection under natural, controlled, and noisy conditions \\
\midrule
\cite{P-n4} & Self-built & 94.17\% & 0.9334 & - & 0.9459 & Proposing Compact Convolutional Transformer (CCT)-based model for automated detection of anthracnose and powdery mildew diseases in mango leaves \\
\midrule
\cite{p4-2023} & Paddy Doctor & 98.74\% & 0.9853 & 0.9786 & 0.9818 & Introducing lesion-aware visual transformer with multi-scale contextual feature extraction and weakly supervised lesion localization \\
\midrule
\cite{p16-2024} & Self-built & 98\% & 0.97 & 0.98 & 0.97 & Introducing Hybrid Pooled Multihead Attention (HPMA) model that improves pest classification by effectively capturing local and global features \\
\midrule
\cite{p31-2023} & Self-built & - & 0.98 & 0.97 & - & Proposing lightweight NanoSegmenter model based on Transformer architecture for high-precision and efficient tomato disease detection, incorporating inverted bottleneck, quantization, and sparse attention techniques \\
\midrule
\cite{p26-2023} & Field-PV, PlantVillage, Syn-PV & - & Per-class & Per-class & - & Proposing Cross-Resolution Transformer (CRFormer) with large-kernel attention and multi-path feed-forward network for accurate grape leaf disease segmentation in complex backgrounds \\
\midrule
\cite{p21-2024} & \makecell[tl]{PlantVillage \\ Cassava leaves \\ Tomato leaves} & \makecell[tl]{99.8\% \\ 84.29\% \\ 99.9\%} & \makecell[tl]{0.9998 \\ 0.8107 \\ 0.9980} & \makecell[tl]{0.9997 \\ 0.8205 \\ 0.9993} & \makecell[tl]{0.9997 \\ 0.7929 \\ 0.9990} & Combining EfficientNet with group-wise Transformer for leaf disease detection \\
\midrule
\cite{p18-2024} & \makecell[tl]{PlantVillage \\ PlantDoc} & \makecell[tl]{99.92\% \\ 75.72\%} & \makecell[tl]{0.996 \\ 0.75} & \makecell[tl]{0.996 \\ 0.74} & \makecell[tl]{0.995 \\ 0.72} & Proposing Leafy\gls*{gan}, a \gls*{gan}-based augmentation framework for synthetic leaf disease data generation, enabling lightweight MobileViT models to achieve high diagnosis accuracy even with limited real data \\
\midrule
\cite{p45-2024} & Potato Leaf Dataset (Kaggle) & 99.33\% & Per-class & - & - & Using ensemble models for potato disease classification \\
\midrule
\cite{p42-2024} & PlantVillage \& Kaggle & Varies by voting ways & Varies by voting ways  & Varies by voting ways & Varies by voting ways & Using ensemble models for leaf disease classification \\
\midrule
\cite{p22-2024} & PlantVillage & Per-class & Per-class & Per-class & Per-class & Using ensemble of \gls*{swint} and residual \gls*{cnn} models \\
\midrule
\cite{p9-2022} & \makecell[tl]{PlantVillage \\ AIChallenger2018 \\ Self-built} & \makecell[tl]{99.84\% \\ 86.83\% \\ 98.78\%} & \makecell[tl]{0.9850 \\ 0.8534 \\ -} & \makecell[tl]{0.9898 \\ 0.8342 \\ -} & \makecell[tl]{0.9865 \\ 0.8539 \\ -} & Proposing a hybrid model (ConvViT) for kiwifruit disease with improved patch embedding and reduced complexity \\
\midrule
\cite{p14-2022} & \makecell[tl]{PlantVillage \\ Custom1 \\ Custom2} & \makecell[tl]{99.62\% \\ 96.58\% \\ 95.97\%} & - & - & - & Proposing a hybrid model of \gls*{cnn} and Transformer to enhance diagnosis in complex scenes \\
\midrule
\cite{p24-2024} & \makecell[tl]{PlantVillage \\ Dhan-Shomadhan} & \makecell[tl]{97.43\% \\ 96.22\%} & \makecell[tl]{0.97 \\ 0.96} & \makecell[tl]{0.96 \\ 0.95} & \makecell[tl]{0.96 \\ 0.95} & Introducing an interpretable hybrid model of lightweight \gls*{cnn} and Transformer for rice leaf disease \\
\midrule
\cite{p50-2023} & \makecell[tl]{PlantVillage \\ Wheat Rust Dataset} & \makecell[tl]{98.81\% \\ 99.86\%} & \makecell[tl]{0.9872 \\ 0.9978} & \makecell[tl]{0.9876 \\ 0.9965} & \makecell[tl]{0.9873 \\ 0.9971} & Proposing hybrid model combining transfer learning and Vision Transformer for deep feature extraction and classification of plant diseases \\
\midrule
\cite{p23-2023} & PlantVillage & 99.8\% & - & - & 0.9931 & Proposing hybrid model combining adaptive Fourier Neural Operators and \glspl*{cnn} to enhance global and local feature extraction for plant leaf disease recognition \\
\midrule
\cite{p40-2024} & Coffee leaf stresses dataset & 99.1\% & - & - & 0.9932 & Proposing a Dual-path \gls*{vit} for efficient and interpretable diagnosis \\
\midrule
\cite{p37-2023} & Citrus datasets & 98.19\% & 0.9839 & 0.9819 & - & Proposing a dual-branch network combining GSDFP (Group Shuffle Depthwise Feature Pyramid) for local multi-scale feature extraction and \gls*{swint} for global context learning for citrus disease classification \\
\midrule
\cite{p28-2024} & PlantPathology & 97.32\% & 0.9733 & 0.9740 & 0.9736 & Introducing DBCoST, a dual-branch architecture that integrates \gls*{cnn} for extracting local features and \gls*{swint} for capturing global information, with a Feature Fusion Module (FFM) to enhance disease identification in apple leaves \\
\midrule
\cite{P-n1}  & \makecell[tl]{Custom \\ CD\&S} & \makecell[tl]{93.01\% \\ 99.69\%} & \makecell[tl]{0.9299 \\ 0.9969} & \makecell[tl]{0.9301 \\ 0.9969} & \makecell[tl]{0.9299 \\ 0.9969} & Proposing the DenseCNNs and Transformer Network (DCTN), featuring a novel multi-head self-attention mechanism for accurate detection of field crop diseases \\
\midrule
\cite{p34-2023} & AIChallenger2018 & 99.00\% & - & - & Per-class & Proposing the TSTC network, a triple-branch Swin Transformer model for simultaneous disease and severity classification, utilizing multitask feature extraction, compact bilinear pooling, and deep supervision to enhance performance and achieve high accuracy on both tasks \\
\midrule
\cite{p25-2024} & AIChallenger2018 & Varies by plant & - & - & - & Proposing LDI-NET, a multi-label network for simultaneous identification of plant type, leaf disease, and severity using a single-branch model, combining \gls*{cnn} and Transformer strengths for feature extraction and employing a multi-label decoder for improved feature utilization \\
\midrule
\cite{p20-2023} & Self-built & 99.10\% & 0.9777 & 0.9847 & 0.9775 & Using Multi-Head Self-Attention (MSA) and a Spatial Convolutional Self-Attention-based Transformer (SCSA-Transformer) for accurate and efficient recognition of multiple strawberry disease classes \\
\midrule
\cite{P-n3} & 5 Public, 1 Self-built & Varies by dataset & Varies by dataset & Varies by dataset & Varies by dataset & Proposing a real-time plant disease identification system using drone-based surveillance, featuring a lightweight \gls*{vit} and \gls*{cnn} fusion model with conditional attention and statistical squeeze-and-excitation \\
\midrule
\cite{pnn-5} & 5 Public & Varies by dataset & Varies by dataset & Varies by dataset & Varies by dataset & Proposing an ultra-lightweight and interpretable hybrid model that combines \gls*{cnn} and \gls*{vit} in a parallel architecture for accurate and efficient plant disease classification, suitable for deployment on resource-constrained IoT devices \\
\midrule
\cite{pnn-6} & 5 Public , 1 Custom  & Varies by dataset & Varies by dataset & Varies by dataset & Varies by dataset & Proposing a novel lightweight hybrid architecture that combines \gls*{cnn} and \gls*{vit} components through a dual-stream token and channel mixer, along with a Bilinear Attention Transformation module for efficient local-global feature extraction \\

\end{longtable}}

\section{Findings and Open Challenges}
\label{findingchallenges}
\subsection{Key Findings}
\paragraph{}
Several important findings have emerged from the review of \glspl*{vit} in precision agriculture. \glspl*{vit} have shown substantial improvements over traditional \glspl*{cnn} in various agricultural tasks, notably in plant disease detection and crop monitoring. Their ability to capture long-range relationships within images enables more accurate identification of plant diseases and pests, even in complex environments. Combining \glspl*{vit} with \glspl*{cnn} in hybrid models has further boosted classification accuracy and robustness, making them a promising approach for agricultural challenges. Furthermore, transfer learning, in which \glspl*{vit} pre-trained on large datasets are fine-tuned for agricultural applications, has proven effective in addressing data limitations and improving model performance without requiring vast amounts of labeled data.
We can summarize the key findings as follows:
\begin{itemize}
    \item \textbf{\glspl*{vit} demonstrate strong performance in plant disease detection:} Many studies have shown that \glspl*{vit} can achieve high accuracy in classifying plant diseases, often outperforming traditional \gls*{cnn} models. Their ability to learn long-range dependencies in images is particularly beneficial for plant disease detection, where symptoms can be distributed across the leaf.
    
    \item \textbf{Self-attention mechanism is a key advantage:} The self-attention mechanism allows \glspl*{vit} to focus on the most important parts of an image without manual feature extraction. This is helpful in detecting small, subtle spots on leaves. \glspl*{vit} can dynamically weigh different regions of the input image, ensuring that significant features receive adequate attention during classification. This helps capture both local and global contextual information. This approach is different from \glspl*{cnn}, which apply convolution operations uniformly across an image and can sometimes miss small but significant features. The ability of \glspl*{vit} to focus on key features helps improve accuracy and reduce the need for extensive pre-processing of images.
    
    \item \textbf{\glspl*{vit} can be optimized for improved performance:} Transfer learning can be effectively used to enhance the performance of \glspl*{vit} for plant disease identification, even with limited data. This involves pre-training the model on a large dataset and then fine-tuning it on a smaller, disease-specific dataset. Attention head pruning can reduce model size and improve inference speed by removing less important attention heads without significant loss of accuracy. Sparse matrix multiplication in self-attention blocks can improve training efficiency and reduce GPU consumption without sacrificing performance. Knowledge distillation can be used to train a smaller, faster model that retains the performance of a larger model, which is especially useful in situations where computational resources are limited.
    
    \item \textbf{Hybrid models are highly effective in improving performance: } Combining \glspl*{vit} and \glspl*{cnn} can leverage the strengths of both architectures. For example, \glspl*{cnn}, by extracting local features, can help \glspl*{vit} see better, allowing the model to capture fine-grained details of the image while maintaining the ability to understand global relationships and context. This ultimately improves performance in tasks such as plant disease detection and image classification.
\end{itemize}
These findings highlight the potential of \glspl*{vit} in precision agriculture and demonstrate their effectiveness in automated plant health monitoring systems, contributing to precision farming. These developments underscore the increasing importance of \glspl*{vit} in revolutionizing precision agriculture, particularly for tasks that demand high accuracy and scalability in large-scale agricultural operations.

\subsection{Open Challenges}
\paragraph{}
Despite the significant progress made with \glspl*{vit} in precision agriculture, several challenges still hinder their wider adoption. A key issue is the absence of inductive biases in \glspl*{vit}. Unlike \glspl*{cnn}, which take advantage of spatial hierarchies and local patterns in images through convolutional filters, \glspl*{vit} do not possess these built-in inductive biases. Consequently, ViTs require large, high-quality datasets to perform well, making them data-hungry. This reliance on vast amounts of data can be a major obstacle in agriculture, where acquiring labeled datasets is both costly and time-consuming. As a result, \glspl*{vit} may struggle to reach optimal performance without substantial data augmentation or pre-training on large-scale datasets like ImageNet.

Another challenge is the high computational cost and resource demands of \glspl*{vit}. The self-attention mechanism that enables \glspl*{vit} to capture long-range dependencies within images is computationally expensive, particularly when dealing with the high-resolution images common in agricultural applications. This leads to high memory and processing power requirements, making it difficult to deploy \glspl*{vit} on resource-limited devices such as drones or edge sensors used in agricultural fields. Additionally, training \glspl*{vit} can take considerably longer than training \gls*{cnn}-based models, which may hinder their use in real-time agricultural applications where fast responses are essential.
Lastly, generalizing across varied agricultural environments presents another significant challenge. While \glspl*{vit} have demonstrated strong performance on specific datasets, they may not be as effective in different climates, crop types, or farming practices. The agricultural landscape is highly diverse, and models trained on data from one region or crop type may fail to adapt to new environments with different lighting conditions, backgrounds, or plant variations. This lack of environmental robustness could limit the practical application of \glspl*{vit} in precision agriculture on a global scale. Therefore, enhancing the ability of \glspl*{vit} to generalize across diverse conditions is a critical area for future research. We can summarize the open challenges as:

\begin{itemize}
    \item \textbf{Need for more comprehensive and diverse datasets:} Many studies use publicly available datasets such as PlantVillage, which may not fully represent the variability of real-world scenarios. Real-world conditions involve a multitude of factors, including diverse plant species, various disease types and stages, and different environmental and lighting conditions, which are often not captured in standard datasets. Datasets also need to account for image variations, including different angles, distances, and image quality. The lack of diverse data can lead to models that are not robust and may not generalize well to new, unseen data. Although some studies have created their own datasets to address this issue, collecting and annotating large-scale, high-quality datasets remains expensive and time-consuming. Table~\ref{tab:tabledata} shows the number of times each dataset has been used.
    
    \item \textbf{Improving the interpretability of models:} While \glspl*{vit} have shown promising results, their internal workings can be less interpretable than those of traditional \glspl*{cnn}. The self-attention mechanisms in \glspl*{vit} enable the model to learn complex relationships between different parts of an image, but it is not always clear why the model makes a particular prediction or which features are most influential in its decision-making process. This lack of interpretability can make it difficult for farmers and agronomists to trust the model or understand how to use it effectively. Some studies address this by using techniques such as Grad-CAM to visualize the areas of an image that are most important for classification. Further research is needed to develop methods that provide deeper insight into the inner workings of \glspl*{vit} and the rationale behind their predictions, such as feature visualization and saliency maps.
    \item \textbf{Reducing computational costs:} Transformer-based models can be computationally expensive, requiring significant computing resources. This cost stems from the self-attention mechanism, whose complexity scales with the input size, making it more difficult to train on larger images and datasets. As a result, deploying these models on resource-constrained devices, such as smartphones or edge devices, can be challenging, particularly in areas with limited access to high-performance computing infrastructure. Some studies propose solutions such as pruning less important attention heads or using sparse matrix multiplication to reduce computational overhead and enhance practicality in real-world applications. Other approaches, including knowledge distillation and hybrid models that combine \glspl*{vit} with \glspl*{cnn}, also aim to improve the efficiency of \glspl*{vit} for deployment. Further research is needed to develop more efficient architectures and training techniques to make \glspl*{vit} more accessible and suitable for practical use in plant disease detection.
    \item \textbf{Addressing the severity of disease:} Most existing studies focus only on detecting the presence or absence of a disease, without considering its severity. However, disease severity can significantly impact crop yield, and determining it is important for developing appropriate treatment plans. More research is needed to develop methods that not only detect the presence of a disease but also assess its severity. This may involve classifying the disease into different stages or using metrics to quantify its extent. Such approaches would require modifying models to output not just a class label but also a measure of disease severity. Additionally, datasets may need to be labeled with more detailed information, including both disease classes and severity levels. Future studies should explore severity assessment methods to support more practical and informed plant disease management strategies.
\end{itemize} 

\captionsetup[longtable]{skip=1em}
\renewcommand{\arraystretch}{1.2}

\begin{longtable}{p{0.2\textwidth}p{0.5\textwidth}S[table-format=2.0]}
\caption{Frequency of dataset usage across the analyzed studies. Datasets created by gathering images from the internet or by combining other datasets are categorized under the 'Custom' category. Datasets created by the authors themselves are categorized under the 'Self-built' category. Datasets that do not have a unique name or DOI link and are simply accessible from Kaggle are categorized under the 'From Kaggle' category.} \label{tab:tabledata} \\
\toprule
\textbf{Dataset} & \textbf{Description} &\textbf{Number of use} \\
\midrule
\endfirsthead

\toprule
\textbf{Dataset} & \textbf{Description} & \textbf{Number of uses} \\
\midrule
\endhead

\midrule
\multicolumn{2}{r}{\textit{Continued on next page}} \\
\endfoot

\bottomrule
\endlastfoot

PlantVillage \cite{plantvillage-dataset} & Contains over 50,000 images of plant leaves across 14 species and 26 diseases & 20 \\
\midrule
From Kaggle & Other publicly available datasets & 14 \\
\midrule
Self-built & Dataset built by the authors & 12 \\
\midrule
Custom & Datasets gathered from search engines, created by combining other datasets, or not easily accessible & 10 \\
\midrule
PlantPathology2020 \cite{plantpathology-dataset} & Contains over 3,000 high-quality real-life images of apple leaf diseases under varying conditions & 4 \\
\midrule
PlantDoc \cite{plantdocdataset} & Contains over 2,000 images across 13 plant species and 17 disease classes & 4 \\
\midrule
CD\&S \cite{CDS} & Contains over 4,000 corn images comprising field images and augmented images & 2 \\
\midrule
Embrapa \cite{Embrapa-dataset} & Contains over 56,000 images of 171 diseases and other disorders affecting 21 plant species & 2 \\
\midrule
PaddyDoctor \cite{paddydoctor-dataset} & Contains over 16,000 annotated paddy leaf images across 13 classes & 1 \\
\midrule
Plant Seedling \cite{plantseedling} & Contains over 900 unique plants belonging to 12 species & 1 \\
\midrule
PlantCLEF2022 \cite{PlantCELF2022} & Contains over 2,000,000 images and 80,000 classes & 1 \\
\midrule
DRLI \cite{drlidataset} & Contains over 4,000 images of healthy and diseased leaves & 1 \\
\midrule
PlantComposite \cite{plantcomposite-dataset} & Contains over 58,000 images of healthy and diseased leaves & 1 \\
\midrule
Citrus Fruits and Leaves dataset \cite{citrusleaf} & Contains over 600 images of citrus leaves and 150 images of citrus fruits, with five classes for each & 1 \\
\midrule
Dhan-Shomadhan \cite{dhandataset} & Contains over 1,100 images of five rice leaf diseases & 1 \\
\midrule
Strawberry Disease dataset \cite{strawberry-dataset} & Contains 2,500 images of seven types of strawberry diseases & 1 \\
\midrule
Maize and Rice dataset \cite{Maize-and-Rice-dataset} & Contains 500 rice images and 466 maize images & 1 \\
\end{longtable}

These challenges highlight that, while \glspl*{vit} hold great potential for plant disease detection, much work remains to ensure that they are robust, accurate, and practical for real-world applications. Addressing these challenges will be crucial for realizing the full potential of \glspl*{vit} in precision agriculture.
\section{Conclusion}
\label{conclusion}
\paragraph{}
The adoption of \glspl*{vit} in agriculture represents a significant departure from conventional \gls*{dl} approaches, offering considerable potential for capturing complex visual patterns without relying on the strong inductive biases inherent in convolutional architectures. While \glspl*{cnn} have historically demonstrated exceptional performance in agricultural applications due to their locality and translation equivariance, \glspl*{vit} present a more adaptable and scalable alternative. Their effectiveness is further enhanced when combined with specialized training strategies and hybrid architectures that strategically reintroduce inductive biases, enabling a balanced trade-off between flexibility and domain-specific robustness. Collectively, these advancements indicate that \glspl*{vit}, whether as standalone models or within hybrid frameworks, are poised to play a pivotal role in the future of agricultural applications.

In this survey, we have traced the evolution of \gls*{vit}-based approaches in agriculture, from vanilla Transformer models to hybrid architectures that leverage the strengths of both \glspl*{cnn} and Transformers. The reviewed literature demonstrates the potential of \glspl*{vit} across a range of agricultural tasks, including disease identification, yield prediction, and precision agriculture, with most models outperforming or complementing traditional methods. Despite these advances, several challenges remain, including data scarcity, high computational requirements, and the need for task-specific model customization. Addressing these challenges is crucial for fully realizing the potential of \glspl*{vit} in practical agricultural applications. As the field progresses, future research should prioritize the enhancement of model efficiency, the adaptation of models to diverse agricultural domains, and the creation of robust, meticulously annotated agricultural datasets.
\bibliographystyle{plainnat}  
\bibliography{references}     

\end{document}